%% file: main.tex
\documentclass{article} 
\usepackage[preprint]{colm2025_conference}

\usepackage{microtype}
\usepackage{hyperref}
\usepackage{url}
\usepackage{booktabs}

\usepackage{lineno}

\definecolor{darkblue}{rgb}{0, 0, 0.5}
\hypersetup{colorlinks=true, citecolor=darkblue, linkcolor=darkblue, urlcolor=darkblue}

\usepackage{algorithm}
\usepackage{comment}
\usepackage[noEnd=True]{algpseudocodex}
\usepackage{algpseudocodex}
\usepackage{multirow}
\usepackage{booktabs}
\usepackage{makecell}
\usepackage{subcaption}
\usepackage{fancybox}
\usepackage{fancyvrb}
\usepackage{multirow}
\usepackage{tabularx}

\newcommand{\op}[1]{\operatorname{#1}}
\usepackage{pifont}  
\usepackage{amssymb} 
\usepackage{booktabs}
\usepackage{listings}
\usepackage{xcolor}
\definecolor{forestgreen}{RGB}{34,139,34}  
\usepackage{colortbl}
\definecolor{lightgray}{gray}{0.9}
\usepackage{mdframed}

\lstdefinelanguage{json}{
    basicstyle=\ttfamily\scriptsize,
    numbers=left,
    numberstyle=\tiny\color{gray},
    stepnumber=1,
    showstringspaces=false,
    breaklines=true,
    frame=single,
    backgroundcolor=\color{gray!10},
    literate=
     *{0}{{{\color{blue}0}}}{1}
      {1}{{{\color{blue}1}}}{1}
      {2}{{{\color{blue}2}}}{1}
      {3}{{{\color{blue}3}}}{1}
      {4}{{{\color{blue}4}}}{1}
      {5}{{{\color{blue}5}}}{1}
      {6}{{{\color{blue}6}}}{1}
      {7}{{{\color{blue}7}}}{1}
      {8}{{{\color{blue}8}}}{1}
      {9}{{{\color{blue}9}}}{1}
      {:}{{{\color{red}:}}}{1}
      {,}{{{\color{red},}}}{1}
      {"}{{{\color{brown}"}}}{1}
}

\DeclareCaptionFont{white}{\color{white}}
\DeclareCaptionFormat{listing}{%
  \parbox{\textwidth}{\colorbox{gray}{\parbox{\textwidth}{#1#2#3}}\vskip-4pt}}
\usepackage[most,skins,theorems]{tcolorbox}

\tcbset{
  aibox/.style={
    width=\linewidth,
    top=8pt,
    bottom=4pt,
    colback=blue!6!white,
    colframe=black,
    colbacktitle=black,
    enhanced,
    center,
    attach boxed title to top left={yshift=-0.1in,xshift=0.15in},
    boxed title style={boxrule=0pt,colframe=white,},
  }
}
\newtcolorbox{AIbox}[2][]{aibox,title=#2,#1}
\definecolor{lightblue}{rgb}{0.22,0.45,0.70}%


\input{packages/header}

\input{packages/math_commands}

\title{RealCritic: Towards Effectiveness-Driven Evaluation of \\Language Model Critiques}

\usepackage{authblk}

\newcommand\nnfootnote[1]{%
  \begin{NoHyper}
  \renewcommand\thefootnote{}\footnote{#1}%
  \addtocounter{footnote}{-1}%
  \end{NoHyper}
}

\author[1,2]{\bf Zhengyang Tang{$^*$}}
\author[1,3]{\bf Ziniu Li{$^*$}}
\author[1,3]{\bf Zhenyang Xiao{$^*$}}
\author[3]{\bf Tian Ding{$^\ddag$}}
\author[1,3]{\bf Ruoyu Sun}
\author[1]{\\ \bf Benyou Wang{$^\ddag$}}
\author[2]{\bf Dayiheng Liu{$^{\ddag}$}}
\author[2]{\bf Fei Huang}
\author[2]{\bf Tianyu Liu}
\author[2]{\bf Bowen Yu}
\author[2]{\bf Junyang Lin}
\affil[1]{The Chinese University of Hong Kong, Shenzhen}
\affil[2]{Qwen Team, Alibaba Inc.}
\affil[3]{Shenzhen Research Institute of Big Data}

\begin{document}

\maketitle


\nnfootnote{$*$ Equal contribution. $\ddag$ Corresponding authors. Emails: zhengyantang@link.cuhk.edu.cn, ziniuli@link.cuhk.edu.cn, kin@stu.pku.edu.cn (Equal contribution); dingtian@sribd.cn, wangbenyou@cuhk.edu.cn, liudayiheng.ldyh@alibaba-inc.com (Corresponding authors).}

\vspace{-0.5cm}

\begin{abstract}
Critiques are important for enhancing the performance of Large Language Models (LLMs), enabling both self-improvement and constructive feedback for others by identifying flaws and suggesting improvements. However, evaluating the critique capabilities of LLMs presents a significant challenge due to the open-ended nature of the task. In this work, we introduce a new benchmark designed to assess the critique capabilities of LLMs. Unlike existing benchmarks, which typically function in an open-loop fashion, our approach employs a closed-loop methodology that evaluates the quality of corrections generated from critiques. Moreover, the benchmark incorporates features such as self-critique, cross-critique, and iterative critique, which are crucial for distinguishing the abilities of advanced reasoning models from more classical ones. We implement this benchmark using eight challenging reasoning tasks. We have several interesting findings. First, despite demonstrating comparable performance in direct chain-of-thought generation, classical LLMs significantly lag behind the advanced reasoning-based model o1-mini across all critique scenarios. Second, in self-critique and iterative critique settings, classical LLMs may even underperform relative to their baseline capabilities.  We hope that this benchmark will serve as a valuable resource to guide future advancements. The code and data are available at \url{https://github.com/tangzhy/RealCritic}.
\end{abstract}

\section{Introduction}

\begin{figure}[h]
    \centering
    \includegraphics[width=1.0\linewidth]{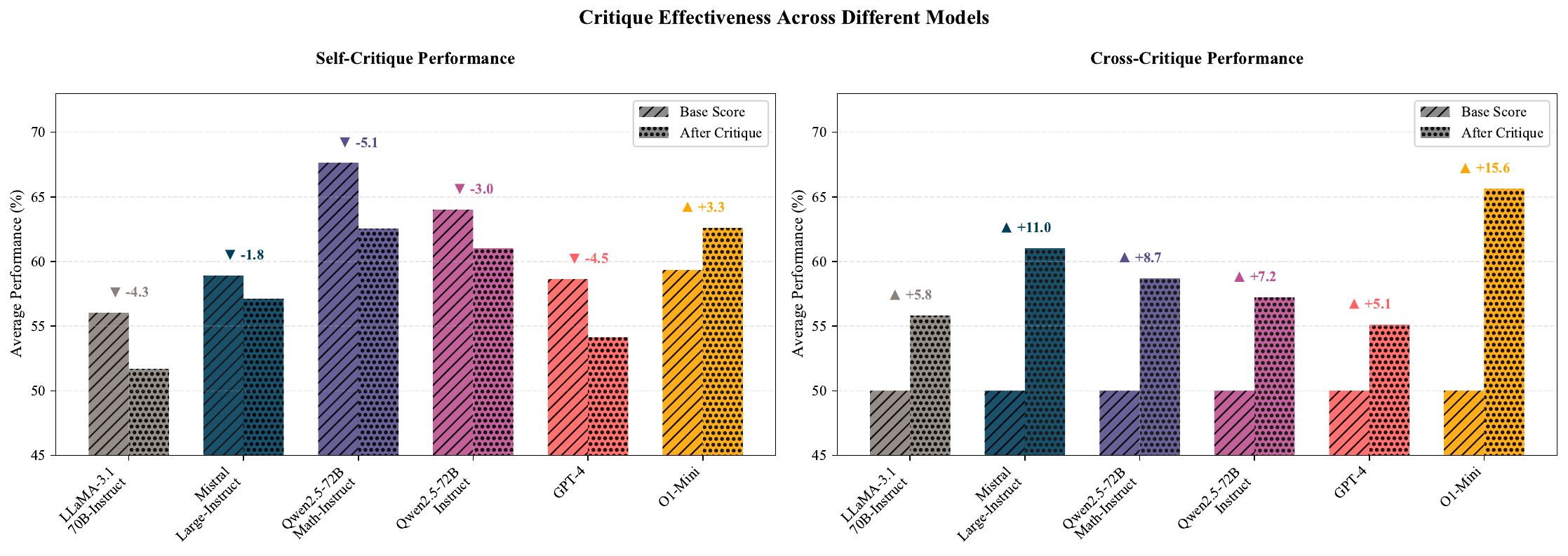}
    \caption{Benchmark results for self-critique and cross-critique abilities of representative LLMs. Our findings reveal a clear seperation between classical LLMs and reasoning-based LLMs: o1-mini is the only model demonstrating improvement in self-critique tasks, while also achieving the most significant gains in cross-critique performance.}
    \label{fig:critique_comparison_paired}
    \vspace{-5pt}
\end{figure}

Recent advancements in Large Language Models (LLMs) \citep{hurst2024gpt4o, dubey2024llama, mistral, qwen2.5} have demonstrated remarkable capabilities across lots of tasks, ranging from natural language understanding to problem-solving. Notably, reasoning-based LLMs have surpassed human expertise in specialized domains such as competition-level mathematical reasoning and code generation \citep{o1mini2024}.  However, as LLMs continue to evolve, further improving their performance has become increasingly challenging. One of the key areas of focus in this regard is the development of \textbf{critique abilities} \citep{mcaleese2024llm}—the capability of LLMs to provide detailed analysis, constructive feedback, and refinement suggestions for solutions generated by other models or even themselves, thereby driving continuous improvement in their outputs.

An important step toward advancing critique abilities is the establishment of high-quality benchmarks for evaluation. Without such benchmarks, it is difficult to objectively measure progress in this area. However, evaluating the quality of critiques is difficult due to the open-ended nature, which lacks a single ground truth answer for definitive judgment.  Furthermore, critiques often serve as intermediate variables within a continuous improvement pipeline, where their ultimate value lies in their ability to refine and enhance subsequent outputs rather than being judged in isolation.

\begin{table*}[t]
    \centering
    \resizebox{\textwidth}{!}{
        \begin{tabular}{@{}l|ccccccc@{}}
        \toprule
        \textbf{Dataset} & 
        \begin{tabular}[c]{@{}c@{}}Output\\ Format\end{tabular} &
        \begin{tabular}[c]{@{}c@{}}Critique\\ Type\end{tabular} &
        \begin{tabular}[c]{@{}c@{}}Evaluation\\ Method\end{tabular} &
        \begin{tabular}[c]{@{}c@{}}Iterative\\ Support\end{tabular} &
        \begin{tabular}[c]{@{}c@{}}Test Data\\ Size\end{tabular} &
        \begin{tabular}[c]{@{}c@{}}Public\\ Release\end{tabular} \\
        \midrule
        CriticBench \citep{luo2023critique} & C+V & Self+Cross & Verdict Matching & \ding{55}  & 0 & \ding{55} \\
        CriticBench \citep{lin2024criticbench}  & C+V & Cross & Verdict Matching & \ding{55}  & 3,825 & \checkmark \\
        CriticEval \citep{lan2024criticeval} & C+S & Cross & Score Judging & \ding{55}  & 3,608 & \checkmark \\
        Shepherd \citep{wang2023shepherd} & C & Cross & GPT-4 Comparison & \ding{55}  & 352 & \ding{55} \\
        Auto-J \citep{li2023generative} & C & Cross & GPT-4 Comparison & \ding{55} & 232 & \checkmark \\
        \midrule
        \textbf{RealCritic} & C+V+Corr & Self+Cross & Correction Matching & \checkmark & 2,093 & \checkmark \\
        \bottomrule
        \end{tabular}
    }
    \caption{Comparison of critique evaluation approaches. C = Critique, V = Verdict, S = Score, Corr = Correction. \checkmark = Yes, \ding{55} = No.}
    \label{tab:benchmark_comparison}
    \vspace{-15pt}
\end{table*}

Significant efforts have been made to benchmark the critique capabilities of LLMs. For instance, CriticBench \citep{luo2023critique,zheng2024critic} requires LLMs to generate critiques accompanied by binary verdicts (C+V) for input solutions, where the verdicts predict whether the solutions are correct or incorrect. The quality of the critiques is then evaluated by comparing these verdicts to ground truth labels, providing a measure of accuracy.  CriticEval \citep{lan2024criticeval} conducts a comprehensive evaluation of LLMs' critique abilities, introducing two key dimensions relevant to our study: feedback and correction. The feedback dimension assesses the quality of the analysis within the critique by assigning a prediction score, while the correction dimension evaluates the quality of revisions based on the critique's analysis. Beyond dedicated benchmarks, other works such as Shepherd \citep{wang2023shepherd} and Auto-J \citep{li2023generative} have proposed using GPT-4 to perform comparative evaluations between different critiques.

Despite these advancements, the key question remains: how can we effectively measure the quality of critiques? This challenge is supported by our preliminary study on CriticBench. Using human evaluators, we find that the benchmark suffers from a high misjudgment 
 rate (about 30\% of low-quality critiques are erroneously classified as high quality; see  \cref{tab:confusion_matrix} in \cref{sec:why_existing_benchmarks_fall_short}). Additionally, we observe that existing benchmarks often lack support for essential critique tasks, such as self-critique and iterative critique, which are believed to be crucial for reasoning-based LLMs \citep{wu2024comparative}.

In this paper,  we propose RealCritic, a new benchmark designed to evaluate critique abilities based on their effectiveness in driving refinement; see \cref{tab:benchmark_comparison} for an overview. Recognizing that a critique is high-quality if it contributes meaningfully to the self-improvement of LLMs, we propose to measure critique quality directly based on the correction it enables. Our approach is closed-loop, drawing inspiration from control theory, where the output (correction) is fed back into the system to assess the effectiveness of the input (critique). In contrast, previous methods for measuring critique quality are often open-loop, evaluating critiques in isolation without considering their downstream impact on solution refinement or treating correction abilities separately from critique abilities. Furthermore, our benchmark introduces advanced evaluation scenarios to assess LLMs' critique capabilities. Specifically, we distinguish between self-critique and cross-critique abilities and incorporate iterative critique, providing a comprehensive framework for evaluating critique effectiveness.

We operationalize RealCritic through 8 challenging tasks: 5 from open-ended mathematical reasoning and 3 from general-domain multiple-choice questions. These tasks are sufficiently challenging and, importantly, have clear ground truth answers for evaluating solution correctness, even though critiques themselves lack ground truth. We evaluate several state-of-the-art models, including LLaMA-3.1-70B-Instruct \citep{dubey2024llama}, Qwen2.5-72B-Instruct \citep{qwen2.5}, Mistral-Large-Instruct-2407 \citep{mistral}, and o1-mini \citep{o1mini2024}. One of our key findings is illustrated in \cref{fig:critique_comparison_paired}, with additional results and analyses detailed in the main text.

To summarize, our contributions are three fold:
\begin{itemize}
\item We introduce a simple yet effective approach to measure critique quality directly based on its impact on solution refinement.
\item We develop a benchmark that supports essential critique functions, including self-critique, cross-critique, and iterative critique, providing a comprehensive evaluation framework tailored to the needs of reasoning-based LLMs.
\item We conduct a comprehensive analysis of existing LLMs' critique abilities using the proposed framework. This analysis highlights the strengths and limitations of current models.
\end{itemize}

\section{Toward Effectiveness-Driven Evaluation}
\label{sec:why_existing_benchmarks_fall_short}

In this section, we discuss strategies for constructing an effective benchmark for evaluating critiques. Evaluating the critique capabilities of LLMs is inherently challenging due to the multifaceted and open-ended nature of the task. A high-quality critique must fulfill several criteria: it should identify flaws or weaknesses in the solution, provide constructive and actionable feedback, and maintain clarity and logical consistency. These requirements complicate the evaluation process, as each sub-part  allows for a wide range of valid responses. While human evaluators may perform well in assessing critique quality, human-based evaluation methods are neither automatic nor scalable.

\begin{figure*}[h]
    \centering
    \includegraphics[width=0.95\linewidth]{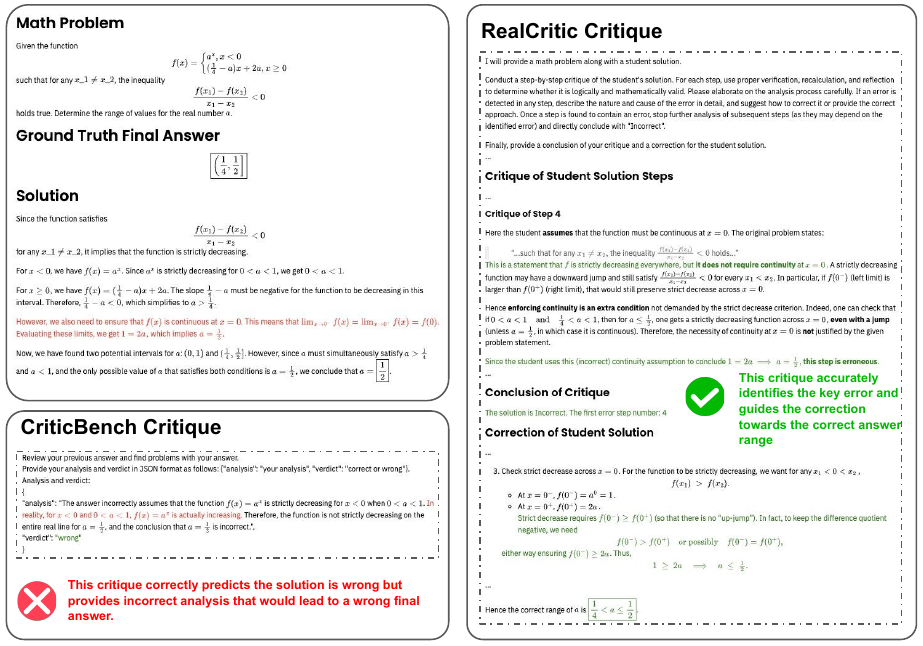}
    \caption{Examples illustrating the limitations of CriticBench \citep{lin2024criticbench}: it incorrectly classifies a low-quality critique as high-quality by relying on the accuracy of predicting the input solution as the metric for critique quality. In contrast, our RealCritic accurately identifies high-quality critiques based on their effectiveness in guiding the generation of improved solutions.}
    \label{fig:criticbench_vs_realcritic}
    \vspace{-10pt}
\end{figure*}

To address these challenges, it is crucial to leverage task-specific properties. Reasoning tasks, for instance, are particularly well-suited for this purpose, as they share a key characteristic: the final generated answer is typically a single, definitive response.  This property can be utilized to facilitate the evaluation of solutions. CriticBench \citep{luo2023critique, lin2024criticbench} leverages this property to evaluate critiques through the following two-step process:
\begin{itemize}
    \item Critic Generation: This benchmark requires LLMs to generate critiques that provide a detailed analysis of a solution. In addition to the analysis, the model must also produce a verdict predicting whether the input solution is correct or incorrect.
    \item Quality Evaluation: The benchmark assesses the quality of the critique based on the correctness of the verdict. If the verdict is correct, the critique is deemed high quality. Conversely, if the verdict is incorrect, the critique is judged as low quality.
\end{itemize}
CriticEval \citep{lan2024criticeval} uses a similar approach in its evaluation framework, particularly in the dimension they define as ``feedback".  We would like to point out that the fundamental premise of this approach is to evaluate critique quality based on the accuracy of predicting the solution's correctness. We argue that this evaluation method may not always be effective, as an LLM might correctly predict the solution's accuracy without providing a constructive analysis or critique. This misalignment stems from the fact that the evaluation approach does not fully capture the intuition that a critique is considered high-quality if it leads to an improved solution, rather than predicting the accuracy of the input solution. A specific example illustrating this limitation is shown in \cref{fig:criticbench_vs_realcritic}. Furthermore, when using human evaluators to provide ground truth assessments of critique quality, we find that CriticBench suffers from a high misjudgment rate, further highlighting its limitations as shown in \cref{tab:confusion_matrix} (see details and more concrete examples in \cref{sec:human_eval_protocol}).

\begin{table}[htbp]
    \centering
    \resizebox{0.45\textwidth}{!}{
        \begin{tabular}{l|cc|c}
            \toprule
            & \multicolumn{2}{c|}{\textbf{Verdict}} & \\
            \textbf{Critique Quality} & Matched & Unmatched & \textbf{Total} \\
            \midrule
            High-quality & 55.3 & 14.6 & 69.9 \\
            \rowcolor{lightgray} Low-quality & 26.0 & 4.1 & 30.1 \\
            \midrule
            \textbf{Total} & 81.3 & 18.7 & 100.0 \\
            \bottomrule
        \end{tabular}
    }
    \caption{Human evaluation results of critique quality on CriticBench~\citep{lin2024criticbench}. All values are in percentages (\%).}
    \label{tab:confusion_matrix}
    \vspace{-10pt}
\end{table}

We believe the aforementioned fragility issue arises from the \textbf{open-loop} nature of the evaluation method employed. Specifically, the quality of the generated critique is not directly linked to its impact on subsequent refinement, leading to potential inconsistencies in evaluation. While this issue is briefly mentioned in CriticEval \citep{lan2024criticeval}, the work does not empirically validate the severity of the issue, as we have done in our study.

In addition to the limitations in evaluation metrics discussed earlier, we also find that existing benchmarks lack comprehensive modes for evaluating critique abilities. First, most benchmarks focus on critiquing input solutions that have been previously collected, operating in a cross-critique mode. However, there is no systematic exploration of another important mode: self-critique, where input solutions are model-specific and dynamically generated. Additionally, existing benchmarks do not include an evaluation mode for multi-round iterative critique, which is essential for examining the long-horizon critique abilities of LLMs and could be particularly crucial for evaluating emergent models such as o1 \citet{o1mini2024}. We plan to address these limitations in the next section by introducing a new benchmark.

\section{Our Evaluation Framework}

\begin{figure*}[h]
    \centering
    \includegraphics[width=0.7\linewidth]{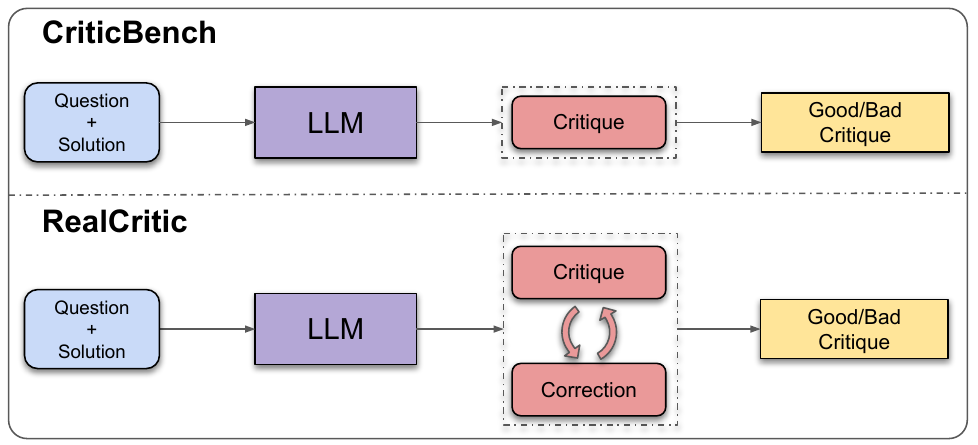}
    \caption{Comparison between the evaluation method used in CriticBench and our framework, RealCritic. Our framework operates in a closed-loop manner by assessing the quality of a critic through the quality of the new solution generated based on its feedback.}
    \label{fig:openloop_vs_closedloop}
    \vspace{-10pt}
\end{figure*}

In this section, we present our evaluation framework for benchmarking the critique abilities of LLMs in mathematical reasoning domains. Our framework is built around three key components: the closed-loop evaluation method described in \cref{sec:closed_loop}, the spectrum of critique abilities outlined in \cref{sec:multi_face_critique}, and the targeted datasets we curate for evaluation, as detailed in \cref{sec:dataset_collection}.

\subsection{From Open-Loop Evaluation to Closed-Loop Evaluation}
\label{sec:closed_loop}

As discussed in \cref{sec:why_existing_benchmarks_fall_short}, we observe that the majority of existing benchmarks rely on open-loop evaluation. This method fails to account for critiques and their ongoing refinement processes, thereby limiting the effectiveness of these benchmarks. To address this limitation, we propose a closed-loop methodology designed to evaluate the model’s critique ability through a \textbf{critique-and-correction} framework. In this framework, the LLM's critiques are not only generated but also tested by applying them to refine the initial solution.

Specifically, we assess the quality of critiques by examining the accuracy of the solution generated after the critique is applied and refined. The key insight driving this approach is that a high-quality critique should naturally lead to a more accurate solution. In other words, the critique is not an isolated task but an integral part of a dynamic correction process that should improve the solution incrementally. This allows us to measure the effectiveness of the critique by evaluating the quality of the final solution. Formally, we define
\begin{align*}
    \op{performance}_{} &= \mathbb{E}_{x \sim \rho}[\mathbb{I}(y_{\op{correction}})], \\
     (y_{\op{critic}}, y_{\op{correction}}) &\sim \mathbb{P}_{\op{Critic-LLM}}(x, s, I_{\op{critic}}),
\end{align*}
where $x$ denotes a question drawn from the distribution $\rho$. The term $\mathbb{P}_{\op{Critic-LLM}}$ represents the distribution over outputs of the (critic) LLM, with $s$ as an initial solution (response) and $I_{\op{critic}}$ as the instruction prompting both  the critique $y_{\op{critic}}$ and correction $y_{\op{correction}}$. The notion $\mathbb{I}$ is the indicator function. A specific example is shown in \cref{fig:criticbench_vs_realcritic}.

We would like to clarify several points. First, while previous works \citep{luo2023critique,lin2024criticbench,lan2024criticeval} also explored measuring the accuracy of the refined solution, this is done in a separate stage and is not used to measure the quality of the critique itself. Second, our approach leverages the quality of the refined solution to measure the quality of the generated critique. However, there are potential shortcuts where models may directly generate a new solution instead of critiquing the original one. To address this, we have carefully tuned the prompt and designed a post-check process to mitigate such behaviors in our evaluations (see Appendix \ref{appendix:post_check}).

\subsection{What to Critique and How to Critique?}
\label{sec:multi_face_critique}

After defining the evaluation criteria, the next step is to design test cases that assess both the model’s critique ability and the underlying critique process. We highlight two key axes often overlooked in existing benchmarks. The first axis distinguishes between self-critique, where the model evaluates its own generated solution, and cross-critique, where it critiques solutions provided by other models. While cross-critique has been widely studied, self-critique remains underexplored, despite its importance for o1-style models that refine outputs through internal feedback loops. The second axis examines whether the critique process is single-round or iterative. Complex tasks often require multi-round critique, where the model refines its feedback across iterations to address errors in initial assessments. This iterative process provides a more comprehensive test of the model’s long-horizon reasoning capabilities. We discuss these points in detail below.

\textbf{Self-Critique Paradigm:} The self-critique paradigm focuses on assessing a model's ability to identify and improve errors in its own generated solutions. Formally, the quality of critique is measured by:
\begin{align*}
\op{performance}_{\op{self}} &= \mathbb{E}_{x \sim \rho}[\mathbb{I} (y_{\op{correction}})], \\
(y_{\op{critic}}, y_{\op{correction}}) &\sim \mathbb{P}_{\op{Critic-LLM}}(x, s, I_{\op{critic}}),  \\
s &\sim  \mathbb{P}_{\op{Critic-LLM}}(x, I_{\op{generation}}). 
\end{align*}
A key feature of this paradigm is that the input solutions for critique are not fixed or pre-collected but are instead model-specific. This paradigm evaluates the model's introspective capabilities, specifically its ability to recognize its own errors, analyze their root causes, and implement effective corrections. However, since the critic and solution generator are the same model, they inherently share the same knowledge base and potential biases, which may limit the diversity and objectivity of the critiques.

\textbf{Cross-Critique Paradigm:} The cross-critique paradigm evaluates a model's ability to analyze and improve answers generated by other models. This approach is widely used in benchmarks, partly because it is straightforward to implement, as input solutions for critique can be pre-collected. This paradigm is crucial for understanding how effectively a model can serve as an external evaluator. Formally, the quality of critique is measured by:
\begin{align*}
\op{performance}_{\op{cross}} &= \mathbb{E}_{x \sim \rho}[\mathbb{I} (y_{\op{correction}})], \\
(y_{\op{critic}}, y_{\op{correction}}) &\sim \mathbb{P}_{\op{Critic-LLM}}(x, s, I_{\op{critic}}),  \\
s &\sim  \mathbb{P}_{\op{LLM}}(x, I_{\op{generation}}), 
\end{align*}
where $\sP_{\op{LLM}}$ may involve any language model distribution, which is distinct from the critic's distribution and could even be a combination of multiple models. This diversity means that the critic must adeptly handle a wide range of solution styles and error types, ensuring versatile evaluation capabilities.

\textbf{Iterative Critique-Correction Paradigm:} Drawing on the success of o1-style models \citep{o1mini2024}, it becomes evident that multi-round self-reflection and iterative refinement play a decisive role in mathematical reasoning. By enabling several cycles of critique-and-correction, we can thoroughly evaluate the model’s continuous improvement across successive iterations, investigate the convergence properties underpinning its error-correction process, and analyze how self-supervision mechanisms influence performance—ultimately shedding light on the model’s capacity for robust and adaptive problem solving.

Formally, we define the iterative process as follows:
\begin{align*}
    \op{performance}_{\op{iterative}}^{t} &= \mathbb{E}_{x \sim \rho}[\mathbb{I} (y_{\op{correction}}^{t})], \\
(y_{\op{critic}}^{t}, y_{\op{correction}}^{t}) &\sim \mathbb{P}_{\op{Critic-LLM}}(x, s, I_{\op{critic}}, y_{\op{critic}}^{1}, y_{\op{correction}}^{1}, \ldots,y_{\op{critic}}^{t-1}, y_{\op{correction}}^{t-1} ),  \\
s &\sim  \mathbb{P}_{\op{LLM}}(x, I_{\op{generation}}), 
\end{align*}

where $t$ denotes the iteration index compared with previous notions.
This formulation highlights the model’s capability to refine its output through repeated self-critique, shedding light on its long-horizon reasoning and error-correction dynamics.

\subsection{Dataset Collection} 
\label{sec:dataset_collection}

\begin{figure}[h]
    \centering
    \includegraphics[width=\linewidth]{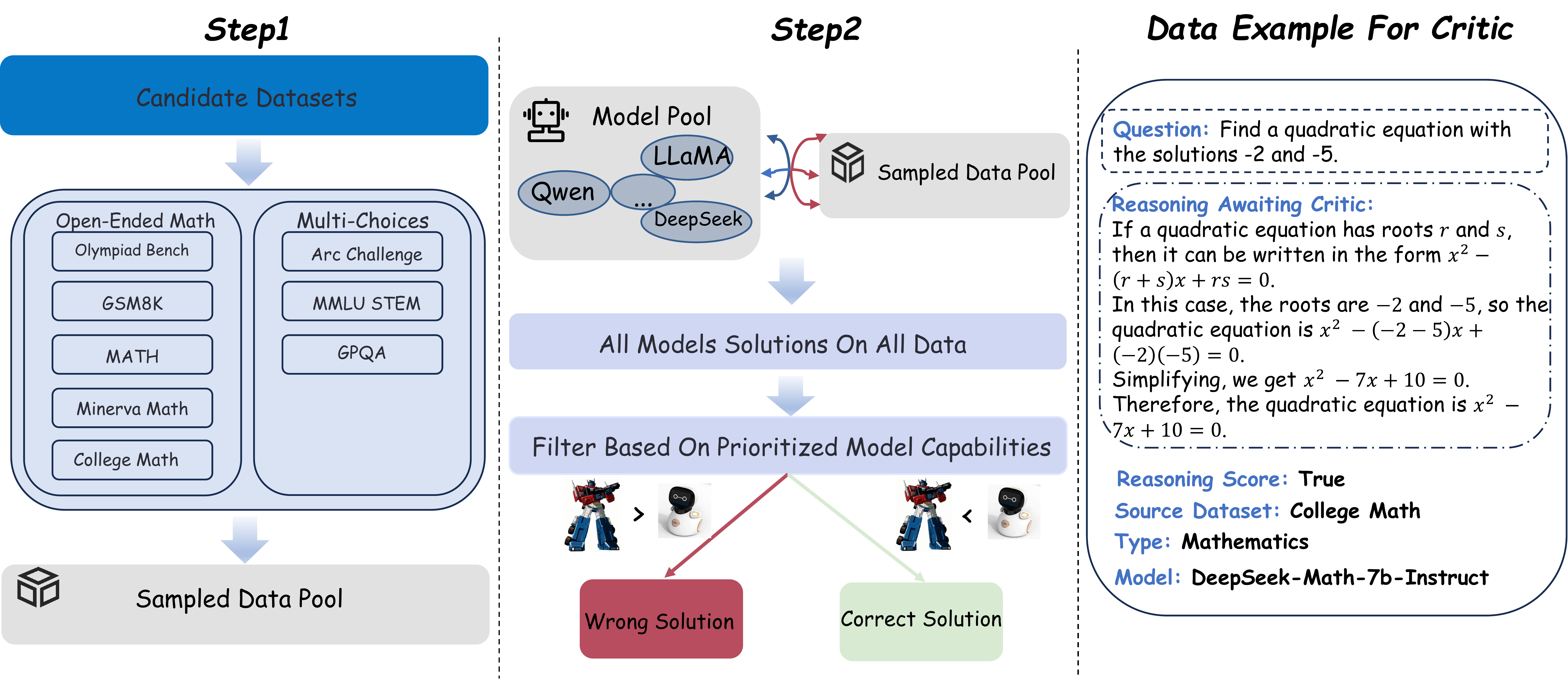}
    \caption{Data collection process for constructing solutions to assess the critique abilities of LLMs.}
    \label{fig:create_benchmark}
    \vspace{-10pt}
\end{figure}

In this section, we present the strategies we propose for constructing high-quality datasets to benchmark critique abilities. An overview of our data collection process is provided in \cref{fig:create_benchmark}. This process involves two key steps: first, collecting questions from a variety of tasks, and second, generating solutions to these questions using multiple models. Both the questions and solutions will be used to evaluate the critique abilities of LLMs.

\textbf{Task Sources.} Our approach began with a comprehensive review of existing evaluation datasets used in prior studies that benchmark well-established models, such as the o1 model \citep{o1mini2024}. We applied a set of selection criteria to filter relevant datasets for our study. Specifically, the datasets had to satisfy the following conditions: they should encompass a variety of difficulty levels, and they must provide explicit final answers to facilitate the verification of solution correctness. After applying these filters, we identified 20 candidate datasets, which encompassed both open-ended math reasoning tasks and general-domain multiple-choice reasoning tasks. For further details on the attributes of these datasets, please refer to  Appendix \ref{appendix:candidate_datasets}. Given the large number of datasets, evaluating all 20 candidates was not feasible within the scope of this study. To ensure a cost-effective and manageable evaluation process, we selected 8 representative datasets that cover a range of difficulty levels and task types.

For open-ended math reasoning tasks, we chose five datasets: GSM8K \citep{cobbe2021gsm8k}, Math \citep{hendrycks2021math}, College Math \citep{collegemath}, Minerva Math \citep{minverva_math}, and Olympiad Bench \citep{he2024olympiadbench}. These datasets span a broad spectrum of mathematical difficulty, from standard high school problems to advanced Olympiad-level challenges, providing a diverse set of tasks that test a model’s reasoning abilities. For general-domain reasoning tasks with multiple-choice questions, we included three datasets: GPQA-diamond \citep{rein2023gpqa}, ARC-Challenge \citep{arc}, and MMLU-STEM \citep{hendrycks2020mmlu}. These datasets assess performance across various domains, with an emphasis on selecting the correct answer from multiple choices. To reduce evaluation costs, we sampled up to 300 items from each dataset, including all items from smaller sets and randomly selecting 300 from larger ones to maintain a diverse yet manageable data pool.

\textbf{Solution Sources.} To generate solutions for the collected questions, we established a model pool primarily composed of open-source models to minimize costs. The pool includes the Qwen2.5 series \citep{qwen2.5} (1.5B, 7B, and 72B), the LLaMA-3.1 series (8B and 72B) \citep{dubey2024llama}, and the Deepseek math-specialized model \citep{deepseekv2} (7B). Given the varying downstream capabilities of these models, we carefully designed a strategy to ensure both the quality and diversity of the generated solutions.

Specifically, we sample each model once on all data points to generate their solutions. The data is then cleaned using the following steps. First, we remove overly long solutions where models fail to stop generating, resulting in repeated outputs until the maximum token limit is reached. Second, we eliminate questions that all models answered correctly, as these are considered too simplistic to yield meaningful insights. Both incorrect and correct solutions are valuable for testing the critique capabilities of LLMs, as they provide distinct challenges for evaluation. To ensure a comprehensive assessment, we apply different strategies for collecting correct and incorrect solutions, which are detailed below.

For incorrect solutions, we prioritize stronger models over weaker ones. Specifically, if multiple models provide incorrect answers to the same question, we select the solution from the strongest model. As illustrated in \cref{tab:wrong_solution_comparison}, when both strong and weak models make errors on the same question, the strong model's solution typically includes more detailed reasoning. This offers the critic model richer information to analyze and identify errors. In contrast, weaker models often produce vague or incomplete incorrect solutions, making it more difficult for the critic model to pinpoint the exact failure points.

For correct solutions, we prioritize weaker models over stronger ones. As shown in \cref{tab:correct_solution_comparison}, strong models generally produce correct solutions with minimal errors in the reasoning process, making them easier for the critic model to evaluate. However, when a weaker model generates a correct solution, it often includes ambiguous or even incorrect intermediate steps, creating a more challenging scenario for the critic model to assess. 

\textbf{Benchmark Format and Usage Protocol.} Finally we detail the key components and usage protocols of our benchmark:

\begin{table*}[h]
\centering
\small 

\begin{tabular}{lcccccccc}
\toprule
& \textbf{ARC} & \textbf{GSM8K} & \textbf{MATH} & \textbf{College} & \textbf{Minerva} & \textbf{GPQA} & \textbf{MMLU} & \textbf{Olympiad} \\
& & & & \textbf{Math} & \textbf{Math} & & \textbf{STEM} & \textbf{Bench} \\
\midrule
Total      & 263 & 273 & 268 & 279 & 261 & 190 & 292 & 267 \\
Correct    & 131 & 136 & 134 & 139 & 130 & 95  & 146 & 133 \\
Wrong      & 132 & 137 & 134 & 140 & 131 & 95  & 146 & 134 \\
\bottomrule
\end{tabular}
\caption{Dataset statistics of input solutions in RealCritic.}
\label{tab:data_format}
\vspace{-10pt}
\end{table*}

\begin{itemize}
    \item \textbf{Dataset Composition}: As shown in Table \ref{tab:data_format}, our benchmark comprises eight distinct datasets spanning various mathematical reasoning domains. For each dataset, we maintain a balanced distribution between correct and incorrect solutions (approximately 1:1 ratio) for cross-critique. We also prepare a separate set of deliberately incorrect solutions for each dataset, with results presented in Appendix \ref{appendix:pure_critique}.
    \item \textbf{Self-Critique Protocol}: The model first generates an initial solution using Chain-of-Thought (CoT) prompting \citep{wei2022chain}, then reviews and corrects its own solution using our critique system prompt.
    \item \textbf{Cross-Critique Protocol}: Given both a question and a pre-collected solution, the model analyzes the solution using the same critique system prompt, identifying potential errors and suggesting specific improvements.
    \item \textbf{Iterative Critique Protocol}: For both self-critique and cross-critique, our benchmark supports multi-turn sessions where the model iteratively refines its analysis until reaching a predefined maximum number of iterations.
    \item \textbf{Concrete Prompts}: Our benchmark utilizes three types of prompts (detailed in Appendix \ref{app:concrete_prompts}): a direct CoT prompt for initial solution generation, a core critique prompt shared by both protocols for systematic analysis and correction, and a follow-up prompt for iterative rounds that enforces fresh analysis. These prompts collectively support our closed-loop evaluation methodology, with concrete examples provided in Appendix \ref{app:concrete_examples}.
\end{itemize}

\section{Experiments}

\subsection{Experimental Setup}

In this section, we conduct experiments using the proposed evaluation benchmark. We provide the experiment configuration below.

\textbf{Models:} We evaluate our framework using a diverse set of state-of-the-art language models, including both open-source models (LLaMA-3.1-70B-Instruct~\citep{dubey2024llama}, Mistral-Large-Instruct-2407~\citep{mistral}, Qwen2.5-72B-Math-Instruct, and Qwen2.5-72B-Instruct~\citep{qwen2.5}) and proprietary models (GPT-4o~\citep{hurst2024gpt4o} and o1-mini~\citep{o1mini2024}).

\textbf{Implementation Details:} For open-source models, we deploy and access them through the vLLM framework to ensure efficient inference. For proprietary models, we utilize their respective official APIs. We maintain consistent generation parameters across all models (except o1-mini which follows its official default settings): temperature = 0.0 (ensuring deterministic outputs), top-p = 1.0, and maximum tokens = 2,048.

\subsection{Results and Analysis}

Our main results are presented in \cref{tab:main_results}, \cref{fig:critique_comparison_paired}, and \cref{fig:iterative_critique}. Depending on the critique abilities being evaluated, we present the concrete analysis in the following sections.

\begin{table*}[h]
\centering
\setlength{\tabcolsep}{4pt}  
\scriptsize  
\begin{tabular}{llccccccccc}
\toprule
\textbf{Model} & \makecell{\textbf{Evaluation}\\\textbf{Mode}} & \textbf{ARC} & \textbf{GSM8K} & \textbf{MATH} & \makecell{\textbf{College}\\\textbf{Math}} & \makecell{\textbf{Minerva}\\\textbf{Math}} & \textbf{GPQA} & \makecell{\textbf{MMLU}\\\textbf{STEM}} & \makecell{\textbf{Olympiad}\\\textbf{Bench}} & \textbf{Avg} \\
\midrule
\multirow{5}{*}{\shortstack[l]{LLaMA-3.1-\\70B-Instruct}} 
& Direct COT & 88.6 & 93.4 & 65.3 & 40.1 & 33.7 & 32.6 & 63.0 & 31.1 & 56.0 \\
& Self-Critique & 87.5 & 91.9 & 59.0 & 37.3 & 33.7 & 22.1 & 53.8 & 28.5 & 51.7 \\
& \textit{$\Delta$(Self-Direct)} & {\color{red} -1.1} & {\color{red} -1.5} & {\color{red} -6.3} & {\color{red} -2.8} & {0.0} & {\color{red} -10.5} & {\color{red} -9.2} & {\color{red} -2.6} & {\color{red} -4.3} \\
& Cross Critique & 82.5 & 88.6 & 64.9 & 49.1 & 42.1 & 35.3 & 42.8 & 40.8 & 55.8 \\
& \textit{$\Delta$(Cross-Base)} & {\color{forestgreen} 32.5} & {\color{forestgreen} 38.6} & {\color{forestgreen} 14.9} & {\color{red} -0.9} & {\color{red} -7.9} & {\color{red} -14.7} & {\color{red} -7.2} & {\color{red} -9.2} & {\color{forestgreen} 5.8} \\
\midrule
\multirow{5}{*}{\shortstack[l]{Mistral-\\Large-Instruct}} 
& Direct COT & 86.7 & 94.5 & 70.1 & 41.6 & 35.6 & 41.1 & 66.8 & 34.5 & 58.9 \\
& Self-Critique & 85.2 & 93.4 & 70.9 & 41.6 & 36.8 & 36.3 & 56.5 & 36.0 & 57.1 \\
& \textit{$\Delta$(Self-Direct)} & {\color{red} -1.5} & {\color{red} -1.1} & {\color{forestgreen} 0.8} & {0.0} & {\color{forestgreen} 1.2} & {\color{red} -4.8} & {\color{red} -10.3} & {\color{forestgreen} 1.5} & {\color{red} -1.8} \\
& Cross Critique & 84.8 & 90.8 & 69.0 & 50.9 & 49.4 & 40.5 & 53.8 & 48.7 & 61.0 \\
& \textit{$\Delta$(Cross-Base)} & {\color{forestgreen} 34.8} & {\color{forestgreen} 40.8} & {\color{forestgreen} 19.0} & {\color{forestgreen} 0.9} & {\color{red} -0.6} & {\color{red} -9.5} & {\color{forestgreen} 3.8} & {\color{red} -1.3} & {\color{forestgreen} 11.0} \\
\midrule
\multirow{5}{*}{\shortstack[l]{Qwen2.5-72B-\\Math-Instruct}} 
& Direct COT & 94.3 & 95.6 & 86.9 & 45.2 & 42.1 & 42.6 & 86.3 & 47.6 & 67.6 \\
& Self-Critique & 93.5 & 95.2 & 85.4 & 45.2 & 41.0 & 41.6 & 50.7 & 47.2 & 62.5 \\
& \textit{$\Delta$(Self-Direct)} & {\color{red} -0.8} & {\color{red} -0.4} & {\color{red} -1.5} & {0.0} & {\color{red} -1.1} & {\color{red} -1.0} & {\color{red} -35.6} & {\color{red} -0.4} & {\color{red} -5.1} \\
& Cross Critique & 72.6 & 86.4 & 73.1 & 52.0 & 50.2 & 43.7 & 38.0 & 53.9 & 58.7 \\
& \textit{$\Delta$(Cross-Base)} & {\color{forestgreen} 22.6} & {\color{forestgreen} 36.4} & {\color{forestgreen} 23.1} & {\color{forestgreen} 2.0} & {\color{forestgreen} 0.2} & {\color{red} -6.3} & {\color{red} -12.0} & {\color{forestgreen} 3.9} & {\color{forestgreen} 8.7} \\
\midrule
\multirow{5}{*}{\shortstack[l]{Qwen2.5-72B\\-Instruct}} 
& Direct COT & 94.3 & 96.7 & 82.5 & 46.2 & 37.9 & 38.9 & 70.9 & 44.2 & 64.0 \\
& Self-Critique & 92.4 & 96.7 & 82.5 & 46.2 & 37.2 & 29.5 & 59.9 & 43.8 & 61.0 \\
& \textit{$\Delta$(Self-Direct)} & {\color{red} -1.9} & {0.0} & {0.0} & {0.0} & {\color{red} -0.7} & {\color{red} -9.4} & {\color{red} -11.0} & {\color{red} -0.4} & {\color{red} -2.9} \\
& Cross Critique & 84.8 & 88.6 & 71.6 & 51.3 & 47.1 & 26.8 & 41.4 & 46.1 & 57.2 \\
& \textit{$\Delta$(Cross-Base)} & {\color{forestgreen} 34.8} & {\color{forestgreen} 38.6} & {\color{forestgreen} 21.6} & {\color{forestgreen} 1.3} & {\color{red} -2.9} & {\color{red} -23.2} & {\color{red} -8.6} & {\color{red} -3.9} & {\color{forestgreen} 7.2} \\
\midrule
\multirow{5}{*}{GPT-4o} 
& Direct COT & 92.4 & 94.9 & 76.5 & 41.6 & 33.0 & 34.2 & 59.9 & 36.3 & 58.6 \\
& Self-Critique & 89.4 & 93.4 & 74.6 & 40.5 & 30.7 & 24.2 & 42.5 & 37.1 & 54.1 \\
& \textit{$\Delta$(Self-Direct)} & {\color{red} -3.0} & {\color{red} -1.5} & {\color{red} -1.9} & {\color{red} -1.1} & {\color{red} -2.3} & {\color{red} -10.0} & {\color{red} -17.4} & {\color{forestgreen} 0.8} & {\color{red} -4.6} \\
& Cross Critique & 82.1 & 84.6 & 68.3 & 49.8 & 43.3 & 30.5 & 37.0 & 45.3 & 55.1 \\
& \textit{$\Delta$(Cross-Base)} & {\color{forestgreen} 32.1} & {\color{forestgreen} 34.6} & {\color{forestgreen} 18.3} & {\color{red} -0.2} & {\color{red} -6.7} & {\color{red} -19.5} & {\color{red} -13.0} & {\color{red} -4.7} & {\color{forestgreen} 5.1} \\
\midrule
\multirow{5}{*}{o1-mini} 
& Direct COT & 68.1 & 89.0 & 81.3 & 20.1 & 28.4 & 56.3 & 81.8 & 49.4 & 59.3 \\
& Self-Critique & 87.5 & 93.8 & 91.4 & 44.1 & 41.0 & 33.2 & 53.8 & 55.8 & 62.6 \\
& \textit{$\Delta$(Self-Direct)} & {\color{forestgreen} 19.4} & {\color{forestgreen} 4.8} & {\color{forestgreen} 10.1} & {\color{forestgreen} 24.0} & {\color{forestgreen} 12.6} & {\color{red} -23.1} & {\color{red} -28.0} & {\color{forestgreen} 6.4} & {\color{forestgreen} 3.3} \\
& Cross Critique & 85.2 & 93.8 & 85.1 & 52.7 & 49.8 & 44.7 & 56.5 & 57.3 & 65.6 \\
& \textit{$\Delta$(Cross-Base)} & {\color{forestgreen} 35.2} & {\color{forestgreen} 43.8} & {\color{forestgreen} 35.1} & {\color{forestgreen} 2.7} & {\color{red} -0.2} & {\color{red} -5.3} & {\color{forestgreen} 6.5} & {\color{forestgreen} 7.3} & {\color{forestgreen} 15.6} \\
\bottomrule
\end{tabular}
\caption{Performance comparison across different evaluation modes. Direct Solution represents the model's ability to solve problems directly. Self-Critique shows performance when the model critiques its own solutions. Cross-Critique indicates performance when critiquing other models' solutions. $\Delta$(Self vs. Direct) shows the improvement from self-critique over direct solution. $\Delta$(Cross vs. Base) shows how much better the model performs compared to the baseline performance (50\%) in cross-model critique tasks. All numbers are in percentages (\%).}
\label{tab:main_results}
\end{table*}

\subsubsection{Self-Critique Performance Analysis}

Our analysis of self-critique capabilities reveals interesting patterns across different models. Most models show varying degrees of performance changes in self-critique compared to their direct CoT solutions, with changes ranging from -5.1\% to +3.3\% on average. This suggests that recognizing and correcting one's own mistakes remains a challenging task even for state-of-the-art language models. However, we notice that o1-mini stands out as a remarkable exception, demonstrating strong self-improvement capabilities with a +3.3\% average improvement over direct solutions. This exceptional performance is particularly evident across multiple tasks: College Math (+24.0\%), ARC (+19.4\%), MATH (+10.1\%), and notably even on the challenging OlympiadBench (+6.4\%). The consistent improvement across diverse mathematical tasks suggests that o1-mini's architecture or training approach might be particularly well-suited for self-reflection tasks.

Furthermore, we observe a concerning pattern across all models in self-critique performance on certain specialized tasks. Models including o1-mini consistently struggle with MMLU-STEM and GPQA tasks, showing significant performance degradation ranging from -10\% to -35.6\%. This pattern is particularly pronounced in Qwen2.5-72B-Math-Instruct's performance on MMLU-STEM (-35.6\%) and o1-mini's performance on MMLU-STEM (-28.0\%). This suggests that self-critique abilities may be more challenging to maintain in complex, specialized domains that require deep subject matter expertise.

\subsubsection{Cross-Critique Performance Analysis}

Cross-critique evaluation reveals a more nuanced picture of models' critique capabilities. The baseline performance is approximately 50\%, as half of the input solutions are correct and half are incorrect. When compared to this baseline, most models show substantial improvements (up to 40\%) on mathematical reasoning tasks. This suggests that models are particularly adept at identifying and correcting errors in basic mathematical reasoning. Similar to the case of self-critique, we observe that o1-mini emerges as the strongest performer in cross-critique tasks. For specialized tasks, models—except for o1-mini—again show degraded performance on average.

\subsubsection{Comparative Analysis of Self-Critique and Cross-Critique}

\begin{figure*}[h]
    \centering
    \includegraphics[width=\linewidth]{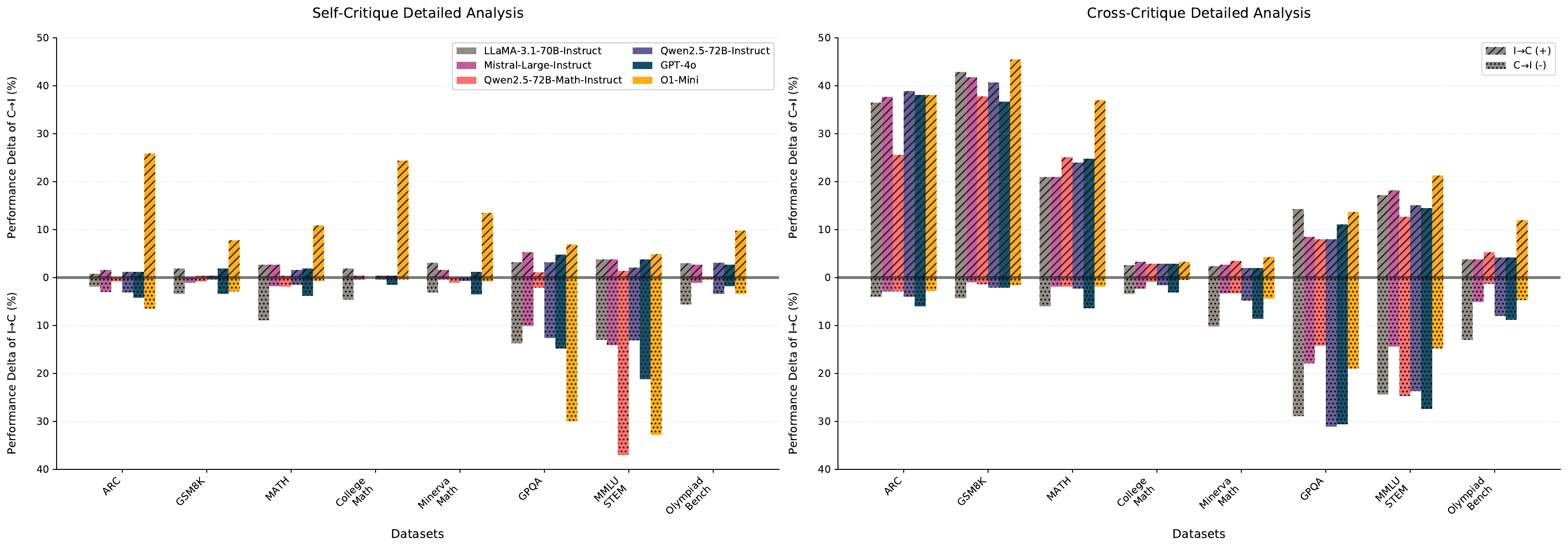}
    \caption{Performance of C→I and I→C in self-critique and cross-critique scenarios. Here, ``C'' denotes correct, ``I'' denotes incorrect, and the arrow indicates the accuracy change from the input solution to the correction after critique.}
    \label{fig:critique_analysis}
\end{figure*}

In this section, we present a more detailed analysis, focusing on two key metrics: I→C, which measures the transformation of incorrect input solutions into correct ones after critique and correction, and C→I, which captures the reversal of originally correct solutions to incorrect ones after critique and correction. These metrics will help us gain deeper insights into the effectiveness of the critique process. The results are reported in \cref{fig:critique_analysis}.

A striking observation is the consistent asymmetry between improvement capability (I→C) and solution maintenance (C→I) across both critique paradigms. Specifically, for self-critique, most models show limited ability to improve incorrect solutions (I→C improvements typically below 5\%), while exhibiting significant degradation when critiquing initially correct solutions (C→I degradations often exceeding -10\%). But, o1-mini stands as a notable exception, demonstrating remarkable improvement capabilities (I→C improvements of 25.85\% on ARC and 24.37\% on College Math) while still struggling with solution maintenance. For cross-critique, we observe substantially higher improvement capabilities across all models (I→C improvements of 30-45\% on fundamental tasks like ARC and GSM8K). However, the degradation issue persists, particularly in specialized domains like GPQA and MMLU-STEM (C→I degradations of -15\% to -30\%). This comparative analysis reveals that while cross-critique generally enables higher potential for improvement, it also carries higher risks of solution degradation. The effectiveness of both critique paradigms appears to be strongly influenced by task complexity and domain specificity, with fundamental mathematical reasoning tasks being more amenable to reliable critique than specialized domain tasks.

\subsubsection{Impact of Iterative Critique}

\begin{figure*}[h]
    \centering
    \includegraphics[width=0.9\linewidth]{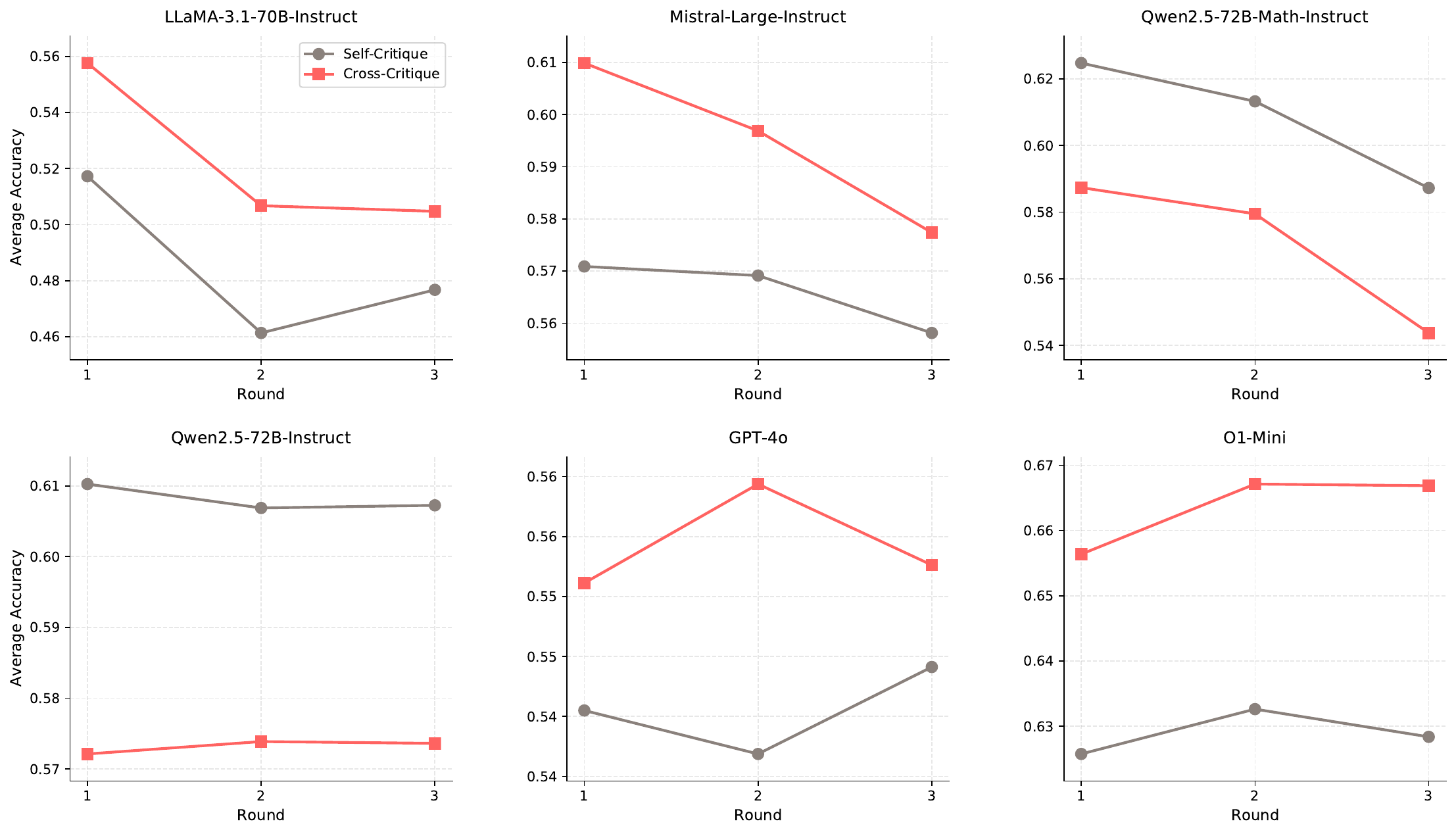}
    \caption{Performance of iterative critique. The performance is averaged over 8 tasks.}
    \label{fig:iterative_critique}
\end{figure*}

In this section, we present the findings from iterative critique experiments, in which models are tasked with providing critiques across multiple rounds. The results are detailed in \cref{fig:iterative_critique}.

We observe distinct trends in how critique effectiveness—measured by performance delta over the base solution—evolves across rounds. LLaMA-3.1-70B-Instruct, Mistral-Large-Instruct, and Qwen2.5-72B-Math-Instruct exhibit a steady decline in effectiveness, while Qwen2.5-72B-Instruct demonstrates remarkable consistency in maintaining its improvement margins. GPT-4o shows an initial increase in cross-critique effectiveness before a decline in self-critique, whereas o1-mini uniquely sustains strong improvement margins throughout iterations, particularly excelling in early rounds. These variations highlight the diverse dynamics of critique effectiveness across models and rounds.

\section{Related Work}

Prior to our work, several efforts have been made to advance the benchmarking of LLMs' critique abilities \citep{luo2023critique, lin2024criticbench, lan2024criticeval}. We have summarized their key features in \cref{tab:benchmark_comparison}. In contrast to these benchmarks, our approach focuses specifically on reasoning tasks and introduces a tightly coupled evaluation framework. This framework validates critique quality through direct correction outcomes and supports diverse critique modes, including self-critique and iterative critique.

Understanding LLMs' ability to self-critique and improve their reasoning has emerged as a critical research direction. \citet{huang2023large} demonstrated that LLMs often struggle with self-correction without external feedback, while \citet{stechly2024self} found that external verification from independent reasoners can significantly improve reasoning accuracy. \citet{wu2024large}'s ProCo framework showed promise in enabling self-correction through minimal prompting and iterative verification. \citet{tyen2023llms} revealed an interesting dynamic: while LLMs may struggle to detect reasoning errors independently, they can effectively correct them when error locations are explicitly identified. Our work builds upon these findings by introducing a novel evaluation framework that examines both self-critique and cross-model critique capabilities.

\section{Conclusion}

In this paper, we present a new benchmark designed to evaluate the critique capabilities of LLMs in reasoning tasks. We investigate three distinct scenarios: self-critique, cross-critique, and iterative critique. Our findings reveal that in nearly all cases, the o1-mini model demonstrates the most impressive performance. While other models may exhibit comparable performance to o1-mini in direct CoT generation, they significantly lag behind in critique abilities. We provide a detailed analysis of these results, offering insights into the underlying factors contributing to these differences. We hope that our findings will inspire and guide future advancements in the development of LLMs.

\bibliography{reference}
\bibliographystyle{named}

\appendix 

\section{Human Evaluation Protocol for Critique Quality}
\label{sec:human_eval_protocol}

We conducted a systematic human evaluation study to establish ground truth labels for critique quality. This section details our evaluation protocol.

\subsection{Data Preparation}
We utilized the evaluation framework from CriticBench \citep{lin2024criticbench}, which provides a comprehensive dataset containing mathematical problems, ground truth solutions, and student solutions. Following their setup, we used 100 problems from their MATH \citep{hendrycks2021math} test set, which already contains mathematical problems, ground truth solutions, and student attempted solutions. We leveraged CriticBench's evaluation code and prompt templates, replacing their default model with Qwen2.5-72B-Instruct\citep{qwen2.5} to generate critiques in their standard JSON format:
    \begin{verbatim}
{
    "analysis": "Detailed analysis of the solution...",
    "verdict": "correct" or "wrong"
}
\end{verbatim}

\subsection{Evaluation Framework}
\label{sec:eval_framework}

Our evaluation framework relies on two key dimensions:
\begin{enumerate}
    \item The correctness of the student's solution (compared to ground truth)
    \item The verdict provided in the critique ("correct" or "wrong")
\end{enumerate}

Based on these dimensions, we developed a comprehensive argument mapping framework to evaluate critique quality, as shown in Table~\ref{tab:argument_mapping}.

\begin{table*}[h]
    \centering
    \resizebox{\textwidth}{!}{
        \begin{tabular}{@{}lll@{}}
            \toprule
            \textbf{Scenario} & \textbf{Condition} & \textbf{Low-Quality Argument} \\
            \midrule
            1 & Solution: Wrong, Verdict: Wrong & 
            \begin{tabular}[t]{@{}l@{}}
                The critique correctly identifies the solution is wrong, but its analysis fails to identify\\
                the true error in the student's reasoning, potentially leading to another incorrect answer.
            \end{tabular} \\
            \midrule
            2 & Solution: Correct, Verdict: Correct & 
            \begin{tabular}[t]{@{}l@{}}
                The critique correctly identifies the solution is correct, but its analysis claims there is\\
                an error in one of the solution steps, suggesting it fails to truly validate the correctness.
            \end{tabular} \\
            \midrule
            3 & Solution: Wrong, Verdict: Correct & 
            \begin{tabular}[t]{@{}l@{}}
                The critique claims the solution is correct, but its analysis contradicts its own verdict\\
                by pointing out an error in one of the solution steps.
            \end{tabular} \\
            \midrule
            4 & Solution: Correct, Verdict: Wrong & 
            \begin{tabular}[t]{@{}l@{}}
                The critique claims the solution is wrong, but its analysis contradicts its own verdict\\
                by confirming every step in the solution is correct.
            \end{tabular} \\
            \bottomrule
        \end{tabular}
    }
    \caption{Argument mapping framework for evaluating critique quality. Each argument describes a specific failure mode that would classify a critique as low-quality. STEM graduate students assessed whether these arguments were true for each critique.}
    \label{tab:argument_mapping}
\end{table*}

\subsection{Evaluation Process}

For each critique evaluation, STEM graduate students were asked to perform a systematic analysis following a standardized procedure:

\begin{enumerate}
    \item First, evaluators thoroughly reviewed the mathematical problem and its ground truth solution to establish a clear understanding of the correct approach and key concepts involved.
    
    \item They then examined the student's attempted solution in detail, comparing it with the ground truth solution to identify any discrepancies or errors.
    
    \item Next, they analyzed the generated critique, paying particular attention to both its verdict ("correct" or "wrong") and the specific reasoning provided in the analysis.
    
    \item Based on the solution's correctness and the critique's verdict, evaluators were presented with the corresponding argument from Table~\ref{tab:argument_mapping}.
    
    \item Finally, evaluators provided: A binary judgment: "True" (indicating the argument applies and the critique is low-quality) or "False" (indicating the argument doesn't apply and the critique is high-quality)
\end{enumerate}

\begin{figure*}[h!]
    \centering
    \includegraphics[width=0.8\linewidth]{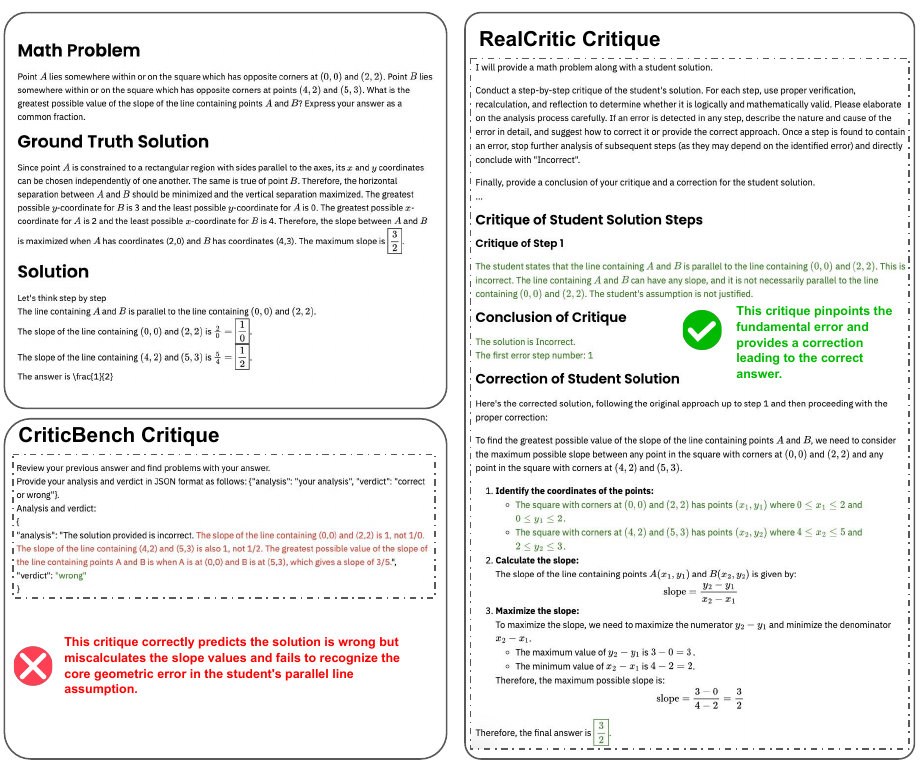}
    \caption{Examples comparing CriticBench and RealCritic.}
    \label{fig:criticbench_vs_realcritic2}
    \vspace{-10pt}
\end{figure*}

\begin{figure*}[h!]
    \centering
    \includegraphics[width=0.8\linewidth]{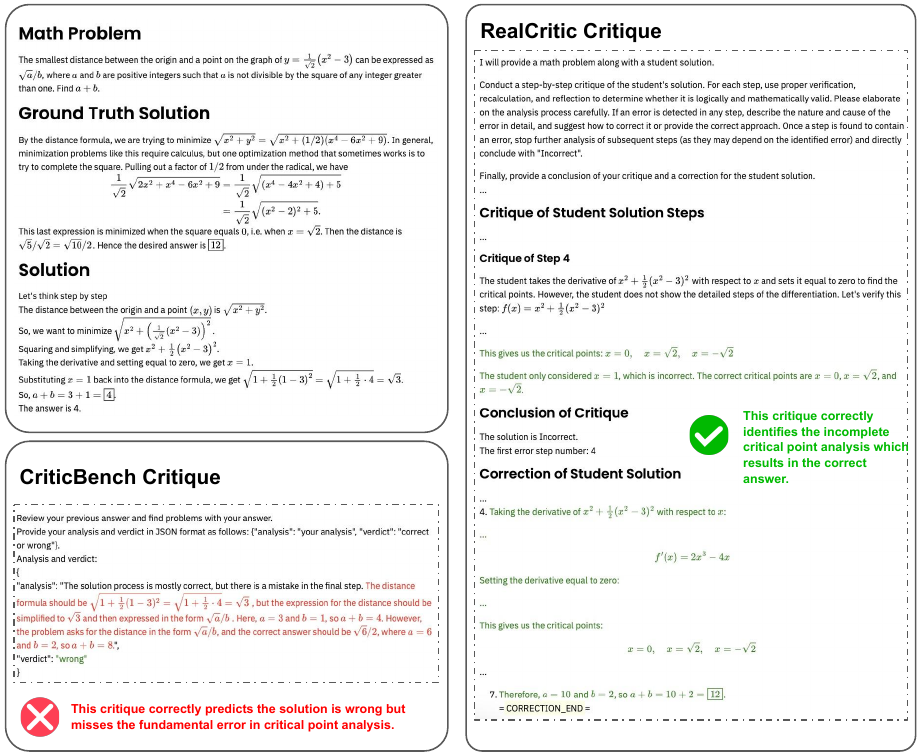}
    \caption{Examples comparing CriticBench and RealCritic.}
    \label{fig:criticbench_vs_realcritic3}
    \vspace{-10pt}
\end{figure*}

\begin{figure*}[h!]
    \centering
    \includegraphics[width=0.8\linewidth]{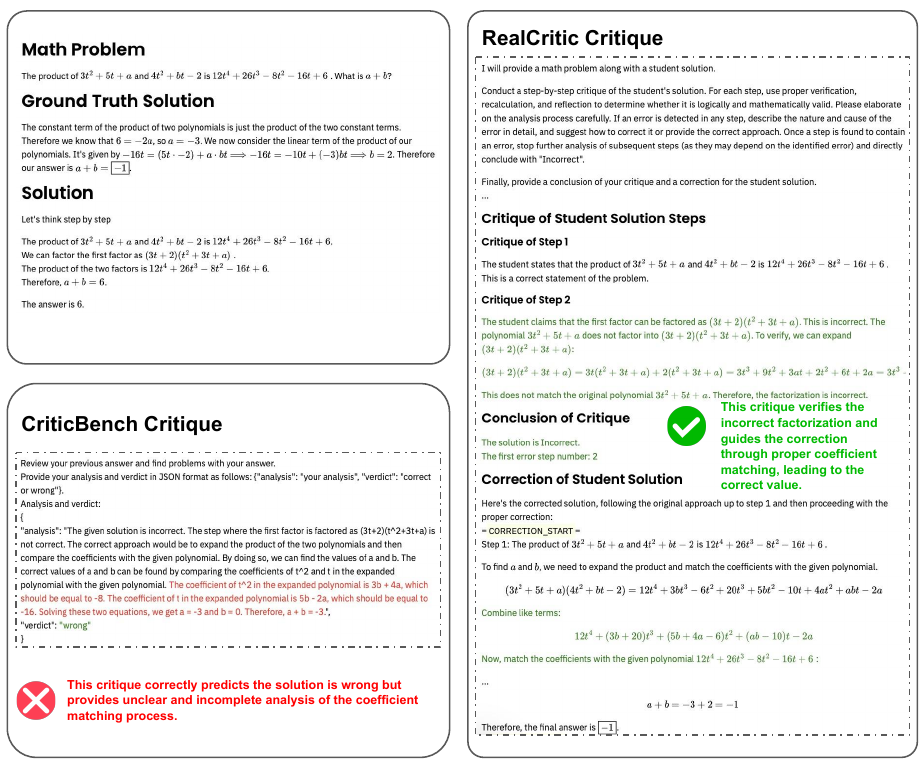}
    \caption{Examples comparing CriticBench and RealCritic.}
    \label{fig:criticbench_vs_realcritic4}
    \vspace{-10pt}
\end{figure*}

The results of this systematic evaluation process were used to create the confusion matrix presented in Table~\ref{tab:confusion_matrix} in the main text, which demonstrates the limitations of verdict-based critique evaluation methods. See Figure \ref{fig:criticbench_vs_realcritic}, \ref{fig:criticbench_vs_realcritic2}, \ref{fig:criticbench_vs_realcritic3}, and \ref{fig:criticbench_vs_realcritic4} for more concrete examples.

\clearpage

\section{Post-Check Mechanism for Critique Quality Assurance}
\label{appendix:post_check}

\begin{table*}[h]
\centering
\resizebox{0.9\linewidth}{!}{
\begin{tabular}{p{\textwidth}}
\toprule
You are a mathematics expert. You will be provided with a question, a solution, and a critique given by a teacher model. The critique will follow this format: If the reasoning is found to be incorrect, first identify the errors in the reasoning to form feedback, and then correct the erroneous reasoning based on the provided feedback.\\
\\
Your task is to evaluate whether the teacher model's critique exhibits any of the following non-compliant behaviors:\\
1. The model does not adhere to the paradigm of critiquing before providing its own solution; instead, it solves the problem first and provides a critique afterward.\\
2. During the critique, the model is capable of identifying errors but fails to correct the flawed reasoning based on the identified mistakes; instead, it circumvents the pointed-out issues in a deceptive manner, providing its own solution.\\
\\
If any of the above non-compliant behaviors are present, the final answer should be \boxed{Unqualified}.\\
If there are no such non-compliant behaviors, the final answer should be \boxed{Qualified}.\\
Please reason step-by-step with your analysis and provide the final answer with \boxed{Unqualified} or \boxed{Qualified}. You can output only these two results. \\
\bottomrule
\end{tabular}
}
\caption{Post-check prompt for verifying the compliance of critique-correction process. This prompt is designed to detect potential shortcuts where models might bypass the proper critique process.}
\label{tab:post_check_prompt}
\end{table*}

To ensure the validity of our closed-loop evaluation framework, we implement a rigorous post-check mechanism to verify that the critique-correction process strictly follows our intended paradigm. This process is crucial for preventing potential shortcuts where models might generate new solutions directly without properly critiquing the original solution.

\subsection{Verification Process}
We employ a carefully designed prompt to evaluate whether each critique adheres to our step-by-step critique-and-correct paradigm in Table \ref{tab:post_check_prompt}.

\subsection{Implementation and Validation}
We utilize Qwen2.5-32B-Instruct as our verification model to evaluate all generated critiques. To validate the reliability of this automated verification process, we conducted a thorough human evaluation study. We randomly sampled 100 critique instances from our dataset for manual review by expert annotators. The results showed a 98\% compliance rate with our critique format requirements, demonstrating the effectiveness of our post-check mechanism in ensuring the quality and validity of the critique-correction process. This verification step is crucial for maintaining the integrity of our evaluation framework, ensuring that the measured critique effectiveness truly reflects the model's ability to identify and correct errors in a systematic manner. Given this high compliance rate, we did not implement post-check as a mandatory step in our evaluation pipeline, as the results indicate that models generally adhere to the intended critique-correction paradigm.

\section{Details On The Attributes Of Candidate Datasets}
\label{appendix:candidate_datasets}

The 20 candidate datasets we selected are: AIME24, AMC23, AQuA~\citep{ling2017program}, Asdiv~\citep{miao2021diverse}, CARP~\citep{zhang2023evaluating}, College Math~\citep{collegemath}, Gaokao2023en~\citep{noauthor_mario-math-reasoninggaokao2023-math-en_nodate}, GSM8K~\citep{cobbe2021gsm8k}, MATH~\citep{hendrycks2021math}, Mawps~\citep{koncel-kedziorski-etal-2016-mawps}, Minerva Math~\citep{minverva_math}, Olympiadbench~\citep{he2024olympiadbench}, BBH~\citep{suzgun2022challenging}, LiveBench~\citep{livebench}, Arc Challenge~\citep{allenai:arc}, GPQA extended, GPQA diamond~\citep{rein2023gpqa}, MMLU-Stem, MMLU-pro, MMLU-redux~\citep{hendrycks2020mmlu}.

The test sets for AIME24 and AMC23 are too small, so we do not consider adding them to RealCritic. The difficulty level of AQuA, Asdiv, CARP, and Mawps is similar to that of GSM8K. We opt to use the classic GSM8K to represent critic performance at this difficulty level. BBH has a wide variety of tasks, which would complicate the evaluation framework, so it is excluded from RealCritic. LiveBench, due to its continuously updated nature, poses high maintenance costs for the future, leading to its exclusion. For the GPQA series, we select the most representative subset, GPQA-diamond. From the MMLU series, we choose MMLU-STEM, which is most closely related to reasoning, for inclusion in RealCritic.

\section{Pure Critique Performance Analysis}
\label{appendix:pure_critique}

\begin{table}[h]
\centering
\setlength{\tabcolsep}{4pt}
\scriptsize
\begin{tabular}{llccccccccc}
\toprule
\textbf{Model} & \makecell{\textbf{Evaluation}\\\textbf{Mode}} & \textbf{ARC} & \textbf{GSM8K} & \textbf{MATH} & \makecell{\textbf{College}\\\textbf{Math}} & \makecell{\textbf{Minerva}\\\textbf{Math}} & \textbf{GPQA} & \makecell{\textbf{MMLU}\\\textbf{STEM}} & \makecell{\textbf{Olympiad}\\\textbf{Bench}} & \textbf{Avg} \\
\midrule
\multirow{2}{*}{\shortstack[l]{LLaMA-3.1-\\70B-Instruct}} 
& Cross Critique & 75.3 & 85.3 & 47.0 & 24.7 & 18.4 & 22.1 & 40.1 & 18.7 & 41.5 \\
& \textit{$\Delta$(Cross-Base)} & {\color{forestgreen} 75.3} & {\color{forestgreen} 85.3} & {\color{forestgreen} 47.0} & {\color{forestgreen} 24.7} & {\color{forestgreen} 18.4} & {\color{forestgreen} 22.1} & {\color{forestgreen} 40.1} & {\color{forestgreen} 18.7} & {\color{forestgreen} 41.5} \\
\midrule
\multirow{2}{*}{\shortstack[l]{Qwen2.5-72B-\\Math-Instruct}} 
& Cross Critique & 54.4 & 74.0 & 51.9 & 28.7 & 16.9 & 18.4 & 37.7 & 18.0 & 37.5 \\
& \textit{$\Delta$(Cross-Base)} & {\color{forestgreen} 54.4} & {\color{forestgreen} 74.0} & {\color{forestgreen} 51.9} & {\color{forestgreen} 28.7} & {\color{forestgreen} 16.9} & {\color{forestgreen} 18.4} & {\color{forestgreen} 37.7} & {\color{forestgreen} 18.0} & {\color{forestgreen} 37.5} \\
\midrule
\multirow{2}{*}{\shortstack[l]{Qwen2.5-72B\\-Instruct}} 
& Cross Critique & 80.6 & 79.5 & 51.1 & 27.6 & 15.7 & 16.3 & 27.4 & 19.5 & 39.7 \\
& \textit{$\Delta$(Cross-Base)} & {\color{forestgreen} 80.6} & {\color{forestgreen} 79.5} & {\color{forestgreen} 51.1} & {\color{forestgreen} 27.6} & {\color{forestgreen} 15.7} & {\color{forestgreen} 16.3} & {\color{forestgreen} 27.4} & {\color{forestgreen} 19.5} & {\color{forestgreen} 39.7} \\
\bottomrule
\end{tabular}
\caption{Performance on deliberately incorrect solutions. All input solutions are incorrect, resulting in a 0\% baseline accuracy. Since there are no correct solutions in the input, the improvement ($\Delta$) equals the absolute performance. All numbers are in percentages (\%).}
\label{tab:pure_critique_results}
\end{table}

To specifically assess models' pure critique and correction capabilities without the influence of correct examples, we conduct additional experiments using deliberately incorrect solutions. Unlike the main experiments where solutions are balanced between correct and incorrect (leading to a 50\% baseline accuracy), here we use exclusively incorrect solutions, resulting in a 0\% baseline accuracy. This setup allows us to cleanly measure how well models can identify and fix errors through critique.

The results show models achieve up to 80\% accuracy on basic tasks like ARC and GSM8K, but performance drops significantly on specialized tasks like Minerva Math and Olympiad Bench (below 20\%), suggesting limited critique abilities on complex mathematical reasoning.

\section{Priority-based Filtering Strategy in Solution Selection}
\label{app:solution_selection}

When we filter out wrong solutions, we should prioritize strong models over weak ones. As shown in Table \ref{tab:wrong_solution_comparison}, when a weak model makes a mistake, it usually provides very little information. This type of solution does not help the critic model perform well and makes evaluation difficult. On the other hand, a strong model typically offers detailed and thorough information, giving the critic model more room to operate effectively.

\begin{table*}[h!]
\centering
\begin{tabular}{p{\textwidth}}
\toprule
\scriptsize
\textbf{Problem.} What is the probability that flipping a fair coin three times produces fewer than two heads? Express your answer as a common fraction.

\vspace{0.5em}
\textbf{Ground Truth Solution.} 
This is the probability of flipping no heads (three tails) or one head (two tails). This is the same as the probability of flipping two heads (one tail) or three heads (no tails), since we can just interchange heads and tails. Since all outcomes are covered, both probabilities add up to 1, so the desired probability is $\boxed{\frac{1}{2}}$.

\vspace{0.5em}
\textbf{Qwen1.5-0.5B-Chat (\textcolor{red}{Wrong Solution})}

The probability that flipping a fair coin three times produces fewer than two heads is given by the binomial coefficient, which is equal to 1/6. Therefore, the answer is \(\frac{1}{6}\).

\vspace{0.5em}
\textbf{Qwen2.5-72B-Math-Instruct (\textcolor{red}{Wrong Solution})}

To determine the probability that flipping a fair coin three times produces fewer than two heads, we first need to consider all possible outcomes when flipping a fair coin three times. Each flip has two possible outcomes: heads (H) or tails (T). Therefore, there are \(2^3 = 8\) possible outcomes in total.

Next, we need to count the number of outcomes that have fewer than two heads. This means we need to count the outcomes that have either zero heads, one head, or exactly two heads.

1. \textbf{Outcomes with zero heads:}

There is only one outcome with zero heads, which is TTT.

2. \textbf{Outcomes with one head:}

There are three outcomes with one head, which are HTT, THT, and TTH.

3. \textbf{Outcomes with exactly two heads:}

There are three outcomes with exactly two heads, which are HHT, HTH, and THH.

Now, we add up the number of outcomes for each case:
\begin{itemize}
    \item Outcomes with zero heads: 1
    \item Outcomes with one head: 3
    \item Outcomes with exactly two heads: 3
\end{itemize}

The total number of outcomes with fewer than two heads is \(1 + 3 + 3 = 7\).

Since there are 8 possible outcomes in total, the probability that flipping a fair coin three times produces fewer than two heads is:

\[\frac{7}{8}\]

Thus, the final answer is:

\[\boxed{\frac{7}{8}}\] \\
\bottomrule
\end{tabular}
\caption{Wrong Solution: Strong Model vs. Weak Model. }
\label{tab:wrong_solution_comparison}
\end{table*}

As shown in Table \ref{tab:correct_solution_comparison}, A capable model generates solutions that, if correct, are not too vague in the process, significantly reducing the critic's burden. In contrast, a less capable model might produce solutions where, even if the final answer is right, the process could have many flaws. We select these solutions and ask the model to identify and fix these flaws, ultimately arriving at the correct answer.

\begin{table*}[h!]
\centering
\begin{tabular}{p{\textwidth}}
\toprule
\scriptsize
\textbf{Problem.} When $\frac{1}{2^{10}}$ is written as a decimal, how many digits are to the right of the decimal point?

\vspace{0.5em}
\textbf{Ground Truth Solution.} 
Multiply numerator and denominator of $\dfrac{1}{2^{10}}$ by $5^{10}$ to see that $\dfrac{1}{2^{10}}$ is equal to $\frac{5^{10}}{10^{10}}$. This implies that the decimal representation of $\dfrac{1}{2^{10}}$ is obtained by moving the decimal point ten places to the left in the decimal representation of $5^{10}$. Therefore, there are $\boxed{10}$ digits to the right of the decimal point in $\dfrac{1}{2^{10}}$.

\vspace{0.5em}
\textbf{Qwen1.5-0.5B-Chat (\textcolor{green}{Answer Correct Solution})}

To find the number of digits to the right of the decimal point when $\frac{1}{2^{10}}$ is written as a decimal, we need to determine the value of $2^{10}$.

The base of the logarithm is $2^{10}$, which is equal to $2^{10 \times 10 \times 10 \times 10 \times 10} = 2^{10 \times 10 \times 10 \times 10 \times 10} = 2^{10 \times 10 \times 10 \times 10} \times 2^1 = 2^{10 \times 10 \times 10 \times 10} \times 2^2 = 2^{10 \times 10 \times 10 \times 10 \times 10} \times 4 = 1024$.

Now, we can find the number of digits to the right of the decimal point by multiplying the base by the number of digits in the base: $1024 \times 10 = 10240$.

So, when $\frac{1}{2^{10}}$ is written as a decimal, there are 10 digits to the right of the decimal point.

\vspace{0.5em}
\textbf{Qwen2.5-72B-Math-Instruct (\textcolor{green}{Answer Correct Solution})}

To determine how many digits are to the right of the decimal point when \(\frac{1}{2^{10}}\) is written as a decimal, we first need to calculate \(2^{10}\).

\[2^{10} = 1024\]

So, \(\frac{1}{2^{10}} = \frac{1}{1024}\). When we write \(\frac{1}{1024}\) as a decimal, it is:

\[\frac{1}{1024} = 0.0009765625\]

Now, we need to count the number of digits to the right of the decimal point. The decimal representation \(0.0009765625\) has 10 digits to the right of the decimal point.

Therefore, the number of digits to the right of the decimal point is \(\boxed{10}\). \\
\bottomrule
\end{tabular}
\caption{Correct Solution: Strong Model vs. Weak Model}
\label{tab:correct_solution_comparison}
\end{table*}

\clearpage
\section{Concrete Prompts}
\label{app:concrete_prompts}
Here we present concrete prompts of RealCritic in Table \ref{tab:prompts} below.

\begin{table*}[h!]
\centering
\resizebox{\linewidth}{!}{
\begin{tabular}{ll}
\toprule
\textbf{Prompt Type} & \textbf{Content} \\
\midrule
Direct CoT & Please reason step by step, and put your final answer within \boxed{}. \\
\midrule
Critique & I will provide a math problem along with a student solution. \\
& \\
& Conduct a step-by-step critique of the student's solution. For each step, use proper verification, recalculation, \\
& and reflection to determine whether it is logically and mathematically valid. Please elaborate on the analysis \\
& process carefully. If an error is detected in any step, describe the nature and cause of the error in detail, and \\
& suggest how to correct it or provide the correct approach. Once a step is found to contain an error, stop further \\
& analysis of subsequent steps (as they may depend on the identified error) and directly conclude with "Incorrect". \\
& \\
& Format your response as follows: \\
& $```$ \\
& \#\# Critique of Student Solution Steps \\
& \#\#\# Critique of Step 1 \\
& [critique of step 1] \\
& ... \\
& \#\#\# Critique of Step n \\
& [critique of step n] \\
& \\
& \#\# Conclusion of Critique \\
& [For correct solution] \\
& The solution is Correct. \\
& The first error step number: -1 \\
& [For incorrect solution] \\
& The solution is Incorrect. \\
& The first error step number: [first step containing error] \\
& \\
& \#\# Correction of Student Solution \\
& [For correct solution] \\
& The student solution is correct and well-reasoned. The final answer is \boxed{[answer]}. \\
& [For incorrect solution] \\
& Here's the corrected solution, following the original approach up to step [error step - 1] and then proceeding \\
& with the proper correction: \\
& $===$CORRECTION START$===$\\
& [Steps 1 to (error step - 1) from original student solution] \\
& [Corrected version of error step] \\
& [Subsequent steps following from the correction to reach the final answer] \\
& Therefore, the final answer is \boxed{[corrected answer]}. \\
& $===$CORRECTION END$===$\\
& $```$ \\
\midrule
Iterative & Are you sure your critic is correct? Please reconsider all the content above. You need to follow the format \\
& of the previous critique, but you are not allowed to copy it !! You must conduct a thorough and sincere \\
& reanalysis on your own! \\
\bottomrule
\end{tabular}
}
\caption{Concrete prompts used in RealCritic for different evaluation scenarios. These prompts are designed to ensure systematic analysis and concrete corrections in our closed-loop evaluation framework.}
\label{tab:prompts}
\end{table*}

\section{Concrete Examples}
\label{app:concrete_examples}

Here we present concrete example of RealCritic in \cref{tab:self_critique_example,tab:cross_critique_example,tab:iterative_critique_example} below.

\begin{table*}[h]
\centering
\begin{tabular}{p{\textwidth}}
\toprule
\scriptsize

\textbf{Step 1: Direct CoT}

\textbf{System Prompt:} Please reason step by step, and put your final answer within \boxed{}.

\textbf{User:}
Let $T=1$. Dennis and Edward each take 48 minutes to mow a lawn, and Shawn takes 24 minutes to mow a lawn. Working together, how many lawns can Dennis, Edward, and Shawn mow in $2 \cdot T$ hours? (For the purposes of this problem, you may assume that after they complete mowing a lawn, they immediately start mowing the next lawn.)

\vspace{0.5em}
\textbf{Assistant:} To determine how many lawns Dennis, Edward, and Shawn can mow together in \(2 \cdot T\) hours (where \(T = 1\)), follow these steps:

...

3. **Calculate Lawns Mowed Individually in 120 Minutes:**
   - **Dennis:** 
     \[
     \frac{120 \text{ minutes}}{48 \text{ minutes/lawn}} = 2.5 \text{ lawns}
     \]
     Since partial lawns aren't counted until completed, Dennis mows **2 lawns**.
   
   - **Edward:**
     \[
     \frac{120 \text{ minutes}}{48 \text{ minutes/lawn}} = 2.5 \text{ lawns}
     \]
     Similarly, Edward mows **2 lawns**.
   
   - **Shawn:**
     \[
     \frac{120 \text{ minutes}}{24 \text{ minutes/lawn}} = 5 \text{ lawns}
     \]
     Shawn mows **5 lawns**.

4. **Total Lawns Mowed Together:**
   \[
   2 \text{ (Dennis)} + 2 \text{ (Edward)} + 5 \text{ (Shawn)} = 9 \text{ lawns}
   \]

**Answer:** 9

\vspace{1em}
\textbf{Step 2: Self-Critique}

\textbf{System Prompt:} I will provide a math problem along with a student solution.

Conduct a step-by-step critique of the student's solution. For each step, use proper verification, recalculation, and reflection to determine whether it is logically and mathematically valid. Please elaborate on the analysis process carefully. If an error is detected in any step, describe the nature and cause of the error in detail, and suggest how to correct it or provide the correct approach. Once a step is found to contain an error, stop further analysis of subsequent steps (as they may depend on the identified error) and directly conclude with "Incorrect".

Finally, provide a conclusion of your critique and a correction for the student solution.

...

\textbf{User:}
The following is the data for your task.

\#\# Math Problem

[copy problem from above]

\#\# Student Solution

[copy solution from above]

\vspace{0.5em}
\textbf{Assistant:} 
\#\# Critique of Student Solution Steps

...

\#\#\# Critique of Step 3

The student correctly calculates the number of lawns each person can mow individually in 120 minutes:

- Dennis: \( \frac{120}{48} = 2.5 \) lawns.

- Edward: \( \frac{120}{48} = 2.5 \) lawns.

- Shawn: \( \frac{120}{24} = 5 \) lawns.

However, the student makes an incorrect assumption by rounding down the partial lawns for Dennis and Edward, resulting in \( 2 \) lawns each. This approach overlooks the fact that all three individuals are mowing simultaneously, and partial progress contributes to the total number of lawns mowed collectively. By rounding down, the student fails to account for the combined efficiency of working together, leading to an underestimation of the total lawns mowed.

\#\# Conclusion of Critique

The solution is Incorrect.
The first error step number: 3

\#\# Correction of Student Solution
Here's the corrected solution, following the original approach up to step 2 and then proceeding with the proper correction:

...

3. **Calculate Combined Mowing Rate:**

   - **Dennis's Rate:** \( \frac{1}{48} \) lawns per minute.
   
   - **Edward's Rate:** \( \frac{1}{48} \) lawns per minute.
   
   - **Shawn's Rate:** \( \frac{1}{24} \) lawns per minute.
   
   **Total Combined Rate:**
   \[
   \frac{1}{48} + \frac{1}{48} + \frac{1}{24} = \frac{1 + 1 + 2}{48} = \frac{4}{48} = \frac{1}{12} \text{ lawns per minute}
   \]

4. **Calculate Total Lawns Mowed in 120 Minutes:**
   \[
   120 \text{ minutes} \times \frac{1}{12} \text{ lawns per minute} = 10 \text{ lawns}
   \]

**Answer:** \boxed{10} \\
\bottomrule
\end{tabular}
\caption{A concrete example of Self-Critique.}
\label{tab:self_critique_example}
\end{table*}

\begin{table*}[h]
\centering
\begin{tabular}{p{\textwidth}}
\toprule
\scriptsize
\textbf{System Prompt:} I will provide a math problem along with a student solution.

Conduct a step-by-step critique of the student's solution. For each step, use proper verification, recalculation, and reflection to determine whether it is logically and mathematically valid. Please elaborate on the analysis process carefully. If an error is detected in any step, describe the nature and cause of the error in detail, and suggest how to correct it or provide the correct approach. Once a step is found to contain an error, stop further analysis of subsequent steps (as they may depend on the identified error) and directly conclude with "Incorrect".

Finally, provide a conclusion of your critique and a correction for the student solution.

...

\textbf{User:}
The following is the data for your task.

\#\# Math Problem

Let $T=37$. A cube has edges of length $T$. Square holes of side length 1 are drilled from the center of each face of the cube through the cube's center and across to the opposite face; the edges of each hole are parallel to the edges of the cube. Compute the surface area of the resulting solid.

\#\# Student Solution

To find the surface area of the resulting solid after drilling square holes through the center of each face of a cube, we need to consider both the original surface area of the cube and the additional surface area created by the holes.

1. **Calculate the original surface area of the cube:**

   The cube has 6 faces, and each face is a square with side length \( T = 37 \). The area of one face is \( 37^2 = 1369 \). Therefore, the original surface area of the cube is:
   \[
   6 \times 1369 = 8214
   \]

2. **Calculate the surface area added by the holes:**

   Each hole is a square with side length 1, and it passes through the entire cube, so it has a length of 37. Each hole has 4 sides, and each side is a rectangle with dimensions \( 1 \times 37 \). The area of one side of a hole is 37, so the total surface area added by one hole is:
   \[
   4 \times 37 = 148
   \]
   Since there are 6 holes (one through each face of the cube), the total additional surface area is:
   \[
   6 \times 148 = 888
   \]

3. **Calculate the surface area of the resulting solid:**

   The surface area of the resulting solid is the sum of the original surface area of the cube and the additional surface area created by the holes:
   \[
   8214 + 888 = 9102
   \]

Thus, the surface area of the resulting solid is:
\[
\boxed{9102}
\]

\vspace{0.5em}
\textbf{Assistant:} 
\#\# Critique of Student Solution Steps

\#\#\# Critique of Step 1

The student correctly calculated the original surface area of the cube. Each face of the cube has an area of \(37^2 = 1369\), and with 6 faces, the total surface area is \(6 \times 1369 = 8214\).

\#\#\# Critique of Step 2

The student incorrectly calculated the additional surface area added by the holes. While each square hole has 4 sides of area \(1 \times 37 = 37\) each, resulting in \(4 \times 37 = 148\) per hole, there are only 3 distinct holes (one along each axis) rather than 6, as drilling through one face automatically accounts for the opposite face. Additionally, the student failed to account for the overlapping areas where the holes intersect at the center of the cube, which leads to double-counting some surface areas. Moreover, the student did not subtract the area removed from the original cube's faces due to the holes.

\#\# Conclusion of Critique

The solution is Incorrect.
The first error step number: 2

\#\# Correction of Student Solution

Here's the corrected solution, following the original approach up to step 1 and then proceeding with the proper correction:

1. **Calculate the original surface area of the cube:**
   The cube has 6 faces, and each face is a square with side length \( T = 37 \). The area of one face is \(37^2 = 1369\). Therefore, the original surface area of the cube is:
   \[
   6 \times 1369 = 8214
   \]

2. **Calculate the surface area removed by the holes:**
   Each square hole removes a \(1 \times 1\) square from each face it penetrates. Since there are 3 pairs of opposite faces, there are 3 distinct square holes. Each hole removes 2 squares (one from each of the two opposite faces), so the total area removed from the cube's original surface is:
   \[
   3 \times 2 \times 1 = 6
   \]
   Therefore, the adjusted original surface area is:
   \[
   8214 - 6 = 8208
   \]

...

5. **Calculate the total surface area of the resulting solid:**
   Add the adjusted original surface area and the net additional surface area from the holes:
   \[
   8208 + 432 = 8640
   \]

Therefore, the surface area of the resulting solid is:
\[
\boxed{8640}
\] \\
\bottomrule
\end{tabular}
\caption{A concrete example of Cross-Critique.}
\label{tab:cross_critique_example}
\end{table*}

\begin{table*}[h]
\centering
\begin{tabular}{p{\textwidth}}
\toprule
\scriptsize
\textbf{System Prompt:} I will provide a math problem along with a student solution.

Conduct a step-by-step critique of the student's solution. For each step, use proper verification, recalculation, and reflection to determine whether it is logically and mathematically valid. Please elaborate on the analysis process carefully. If an error is detected in any step, describe the nature and cause of the error in detail, and suggest how to correct it or provide the correct approach. Once a step is found to contain an error, stop further analysis of subsequent steps (as they may depend on the identified error) and directly conclude with "Incorrect".

Finally, provide a conclusion of your critique and a correction for the student solution.

...

\textbf{User:}
The following is the data for your task.

\#\# Math Problem

The equation

$$
(x-1)(x-2) \cdots(x-2016)=(x-1)(x-2) \cdots(x-2016)
$$

is written on the board. One tries to erase some linear factors from both sides so that each side still has at least one factor, and the resulting equation has no real roots. Find the least number of linear factors one needs to erase to achieve this.

\#\# Student Solution

To solve this problem, we need to manipulate the given equation by erasing some linear factors from both sides in such a way that the resulting equation has no real roots. The original equation is:

$$
(x-1)(x-2) \cdots(x-2016)=(x-1)(x-2) \cdots(x-2016)
$$

...

The least number of linear factors one needs to erase to ensure the resulting equation has no real roots is **1008**.

**Answer:** 1008

\vspace{0.5em}
\textbf{Assistant:} 
\#\# Critique of Student Solution Steps

...

\#\#\# Critique of Step 2

...

Due to these issues, Step 2 contains critical errors in both reasoning and application of mathematical principles.

\#\# Conclusion of Critique

The solution is Incorrect.
The first error step number: 2

\#\# Correction of Student Solution

Here's the corrected solution, following the original approach up to step 1 and then proceeding with the proper correction:

To solve this problem, we need to manipulate the given equation by erasing some linear factors from both sides in such a way that the resulting equation has no real roots. The original equation is:

$$
(x-1)(x-2) \cdots (x-2016) = (x-1)(x-2) \cdots (x-2016)
$$

...

The least number of linear factors one needs to erase to ensure the resulting equation has no real roots is **2**.

**Answer:** 2 

\textbf{User:} Are you sure your critic is correct? Please reconsider all the content above. You need to follow the format of the previous critique, but you are not allowed to copy it !! You must conduct a thorough and sincere reanalysis on your own!

\textbf{Assistant:} \#\# Critique of Student Solution Steps

...

\#\#\# Critique of Step 2

...

Due to these critical flaws, Step 2 does not reliably advance the solution and introduces incorrect assumptions about the nature of \( P \) and \( Q \).

\#\# Conclusion of Critique

The solution is Incorrect.
The first error step number: 2

\#\# Correction of Student Solution

Here's the corrected solution, following the original approach up to step 1 and then proceeding with the proper correction:

To solve this problem, we need to manipulate the given equation by erasing some linear factors from both sides in such a way that the resulting equation has no real roots. The original equation is:

$$
(x - 1)(x - 2) \cdots (x - 2016) = (x - 1)(x - 2) \cdots (x - 2016)
$$

...

The least number of linear factors one needs to erase to ensure the resulting equation has no real roots is **2016**.

**Answer:** 2016

\\
\bottomrule
\end{tabular}
\caption{A concrete example of Iterative-Critique.}
\label{tab:iterative_critique_example}
\end{table*}

\end{document}

%% file: packages/header.tex
\usepackage{url}            
\usepackage{booktabs}       
\usepackage{multirow}    
\usepackage{amsfonts}       
\usepackage{nicefrac}       
\usepackage{microtype}      
\usepackage{natbib}
\usepackage{enumerate}
\usepackage{hhline}
\usepackage{makecell}
\usepackage{pifont}

\usepackage{graphicx} 
\usepackage{amsmath}
\usepackage{amsthm}
\usepackage{amssymb}
\usepackage{tikz}
\usepackage{xcolor}
\usetikzlibrary{arrows}

\allowdisplaybreaks

\usepackage{mathrsfs}

\usepackage{hyperref}
\usepackage{bm}

\allowdisplaybreaks

\usepackage[capitalize,noabbrev]{cleveref}
\crefname{thm}{Theorem}{Theorems}
\crefname{lem}{Lemma}{Lemmas}
\crefname{cor}{Corollary}{Corollaries}
\crefname{prop}{Proposition}{Propositions}
\crefname{asmp}{Assumption}{Assumptions}
\crefname{defn}{Definition}{Definitions}
\crefname{oracle}{Oracle}{Oracles}
\crefname{fact}{Fact}{Facts}
\crefname{conj}{Conjecture}{Conjectures}
\crefname{rem}{Remark}{Remarks}
\crefname{example}{Example}{Examples}
\crefname{condition}{Condition}{Conditions}
\crefname{exercise}{Exercise}{Exercises}
\crefname{algorithm}{Algorithm}{Algorithms}
\crefname{table}{Table}{Tables}
\crefname{figure}{Figure}{Figures}
\crefname{section}{Section}{Sections}
\crefname{subsection}{Section}{Sections}
\crefname{appendix}{Appendix}{Appendices}
\crefname{message}{Message}{Messages}

\definecolor{red}{rgb}{1, 0, 0}

\definecolor{green}{rgb}{0, 1, 0}

\definecolor{blue}{rgb}{0, 0, 1}

\definecolor{orange}{rgb}{1, 0.4, 0.0}